%% file: main.tex
\documentclass{ecai} 

\usepackage{latexsym}
\usepackage{amssymb}
\usepackage{amsmath}
\usepackage{amsthm}
\usepackage{booktabs}
\usepackage{enumitem}
\usepackage{graphicx}
\usepackage{color}

\newcommand{\cmark}{\ding{51}}%
\newcommand{\xmark}{\ding{55}}%

\usepackage{comment}
\usepackage{multirow}
\usepackage{float}
\usepackage{subcaption}

\usepackage{pifont}     

\newcommand{\BibTeX}{B\kern-.05em{\sc i\kern-.025em b}\kern-.08em\TeX}

\begin{document}


\begin{frontmatter}


\paperid{} 


\title{FedVLM: Scalable Personalized Vision-Language Models through Federated Learning}


\author[A]{\fnms{Arkajyoti}~\snm{Mitra}\thanks{Corresponding Author. Email: axm3158@mavs.uta.edu.}} 
\author[A]{\fnms{Afia}~\snm{Anjum}}
\author[A]{\fnms{Paul}~\snm{Agbaje}} 
\author[B]{\fnms{Mert}~\snm{Pesé}} 
\author[A]{\fnms{Habeeb}~\snm{Olufowobi}} 

\address[A]{University of Texas at Arlington}
\address[B]{Clemson University}




\input{0_abstract}

\end{frontmatter}

\begin{figure*}[ht]
  \centering
  \begin{subfigure}[b]{0.66\textwidth}
    \centering
    \includegraphics[width=\textwidth]{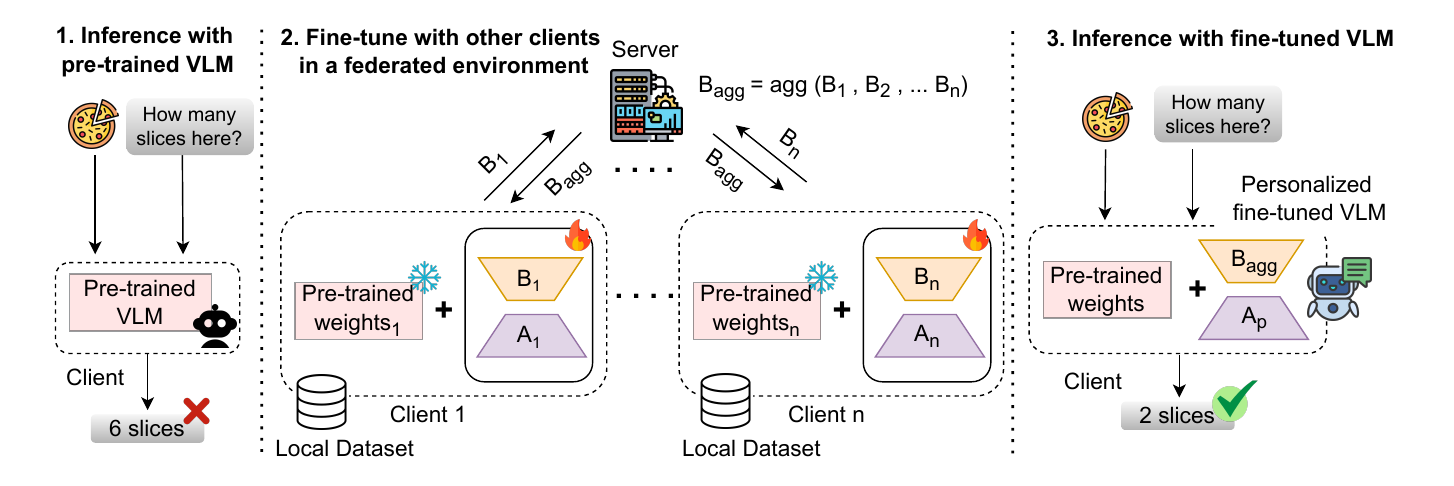}
      \caption{FedVLM pipeline.}
  \end{subfigure}
  \hfill
  \begin{subfigure}[b]{0.31\textwidth}
    \centering
    \includegraphics[width=\textwidth]{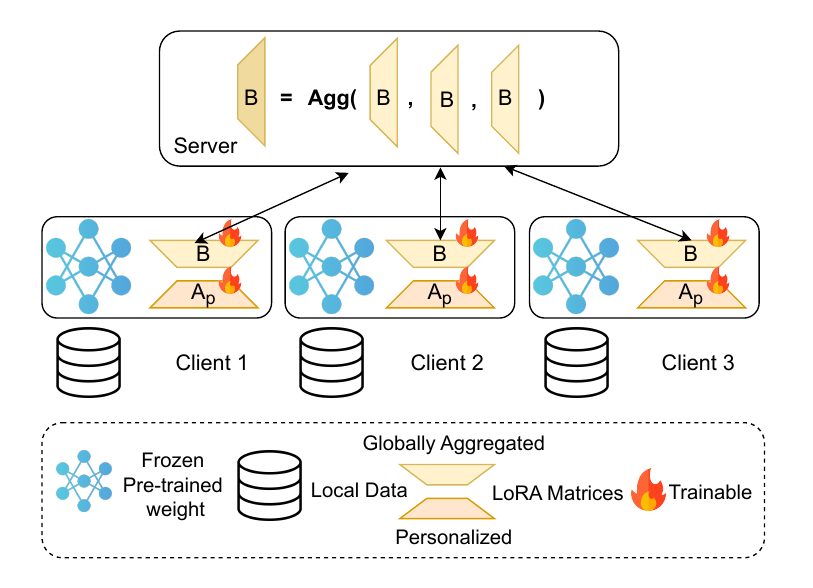}
    \caption{Aggregation process with pLoRA.}
  \end{subfigure}

  \vspace{3mm}
  \caption{\textbf{Overview of FedVLM}: (a) The process begins with inference using a pre-trained VLM, which struggles with task-specific adaptations. Clients then fine-tune the model in FL setting while keeping the pre-trained weights frozen and training pLoRA matrices locally. The pLoRA matrix $B$ are aggregated at the server and redistributed. The final personalized fine-tuned VLM improves performance while maintaining efficiency. (b) The server aggregates LoRA matrices $B$ from multiple clients, distributing globally aggregated parameters while allowing clients to retain personalized components $A_p$. This approach enables efficient adaptation, reducing communication costs while maintaining privacy and personalization. }\label{fig:fedvlm}
  \vspace{6mm}
\end{figure*}

\input{1_intro}

\input{2_background}

\input{3_methodology}

\input{4_experimental_setup}

\input{5_conclusion}






\bibliography{main}

\end{document}

%% file: 0_abstract.tex
\begin{abstract}
     Vision-language models (VLMs) demonstrate impressive zero-shot and few-shot learning capabilities, making them essential for several downstream tasks. However, fine-tuning these models at scale remains challenging, particularly in federated environments where data is decentralized and non-iid across clients. Existing parameter-efficient tuning methods like LoRA (Low-Rank Adaptation) reduce computational overhead but struggle with heterogeneous client data, leading to suboptimal generalization.
     To address these challenges, we propose FedVLM, a federated LoRA fine-tuning framework that enables decentralized adaptation of VLMs while preserving model privacy and reducing reliance on centralized training. To further tackle data heterogeneity, we introduce personalized LoRA (pLoRA) which dynamically adapts LoRA parameters to each client’s unique data distribution, significantly improving local adaptation while maintaining global model aggregation. Experiments on the RLAIF-V dataset show that pLoRA improves client-specific performance by $24.5\%$ over standard LoRA, demonstrating superior adaptation in non-iid settings.
     FedVLM provides a scalable and efficient solution for fine-tuning VLMs in federated settings, advancing personalized adaptation in distributed learning scenarios.
\end{abstract}

%% file: 1_intro.tex
\section{Introduction}

Vision-language models (VLMs) are transformer-based generative models capable of processing multimodal inputs---such as texts and images---and auto-regressively generate outputs. These models have demonstrated remarkable capabilities across various tasks, including classification (e.g., object detection), generation (e.g., image description), and comprehension (e.g., visual question answering)~\cite{saha2024improved}.

Before VLMs, vision-related tasks were typically addressed through modular pipelines, where detection, segmentation, and planning were executed sequentially~\cite{kemsaram2019integrated}. For example, in autonomous driving (AD), lane detection and drivable space segmentation enable vehicles to maintain lane position, while traffic sign recognition aids in decision-making. However, these fragmented approaches introduce latency, increase computational costs, and limit adaptability to novel scenarios. In contrast, VLMs offer end-to-end reasoning capabilities, enabling zero-shot and few-shot inference with minimal task-specific supervision~\cite{li2023blip}.

VLMs have the potential to transform multiple domains, including autonomous driving, where they can analyze real-time imagery for rapid decision-making~\cite{hu2023gaia}; medicine, where they assist in pathology image classification~\cite{huang2023visual}; and law, where they simplify complex legal documents for non-experts~\cite{li2024enhancing}. Despite these advantages, adapting VLMs to specific downstream tasks remains a significant challenge, particularly in decentralized environments where privacy, bandwidth, and resource constraints are critical.

While VLMs benefit from extensive pre-training, fine-tuning them for domain-specific tasks requires significant computational resources and labeled data. Centralized fine-tuning approaches, which aggregate all client data on a central server, are impractical in many real-world settings due to privacy concerns, communication overhead, and the computational limitations of edge devices. On-device fine-tuning of large models is also challenging due to memory constraints, increased inference latency, and high storage costs. This leads to a critical question: 

\noindent
\begin{center}
 \textbf{How can VLMs be efficiently and securely adapted in decentralized environments while addressing client-specific data heterogeneity?}   
\end{center}

To address these challenges, we propose FedVLM, a novel federated learning (FL) framework designed for efficient, privacy-preserving adaptation of VLMs across decentralized clients. To the best of our knowledge, using decentralized training methods like FL for VLMs is still under-explored. FL~\cite{mcmahan2017communication} enables collaborative model training without sharing raw data, preserving privacy while reducing reliance on centralized fine-tuning. However, standard FL suffers under non-iid data settings, as global aggregation can dilute client-specific features.
To address this, we introduce personalized LoRA (pLoRA), a lightweight adaptation technique that dynamically tailors LoRA parameters to each client’s unique data distribution. Unlike standard LoRA, which applies a shared low-rank adaptation across all clients, pLoRA fine-tunes VLMs in a client-specific manner, significantly improving personalization in non-iid federated environments. Our FedVLM framework with pLoRA outperforms state-of-the-art (SOTA) LoRA-based fine-tuning methods in federated settings, demonstrating substantial improvements in personalization and efficiency.

Our key contributions are as follows:
\begin{itemize}
    \item We propose FedVLM, a federated learning framework for VLMs that supports decentralized adaptation under privacy and resource constraints.
    \item We introduce pLoRA, a novel personalization technique that selectively personalize LoRA matrices, enhancing VLM performance in heterogeneous federated environments. 
    \item We conduct extensive empirical evaluation of FedVLM in both iid and non-iid settings, demonstrating its superiority over existing LoRA-based fine-tuning methods.
\end{itemize}

%% file: 2_background.tex
\section{Background}

\subsection{Vision Language Models}
VLMs are designed to interpret and describe objects in images based on textual query prompts. These models, often containing billions of parameters, require large-scale, high-quality datasets of image-text pairs to capture complex visual-textual relationships. However, fine-tuning VLMs for specialized tasks remains a challenge due to the computational cost and privacy concerns associated with centralized data collection.

Most VLMs utilize an encoder-decoder architecture, where separate encoders independently process visual and textual inputs before forming a unified representation~\cite{li2022blip}. The visual encoder, such as OpenCLIP ViT-G, extracts feature embeddings from images, while the text encoder converts queries into corresponding textual embeddings. Some models adopt joint encoders to learn both modalities simultaneously~\cite{li2021align}, potentially improving efficiency but at the cost of increased computational complexity. The choice between separate and joint encoders introduces trade-offs in terms of model accuracy, inference speed, and suitability for deployment on resource-constrained devices.

With the growing need for on-device fine-tuning in settings, such as edge systems, there is increasing demand for smaller, more efficient VLMs that can adapt to diverse environments while maintaining strong generalization. Recent models such as Tiny-LLaVA~\cite{zhou2024tinyllava} and the Phi family~\cite{textbooks2} demonstrate the feasibility of deploying lightweight VLMs in real-time applications. However, most of these models still rely on centralized fine-tuning, limiting their adaptability to personalized tasks and increasing privacy risks.

To address these limitations, FL offers a promising solution for distributed VLM fine-tuning by enabling models to learn from decentralized data without requiring raw data to be shared across devices. FL reduces communication overhead and enhances privacy while maintaining adaptability in dynamic, real-world settings. Despite the FL success in training NLP models, its application to multi-modal VLMs remains an open challenge.
In response, we propose leveraging FL for VLM fine-tuning, introducing efficient techniques to maintain scalability, enhance personalization, and preserve generalization in dynamic, real-world settings.

\subsection{VLM Pre-Training and Fine Tuning}

Most VLMs employ a transformer-based architecture that use attention mechanisms to support reasoning and zero-shot inference~\cite{radford2021learning}. Transformers process input sequences in parallel using self-attention mechanisms, making them highly efficient for handling large-scale datasets and long sequences~\cite{vaswani2017attention}. When provided with sufficient data and computational resources, these architectures can learn rich, generalizable feature representations. However, scaling these models poses significant challenges, as both training and inference become increasingly resource-intensive with model size.

To mitigate these challenges, parameter-efficient fine-tuning (PEFT) methods have emerged as a promising approach, allowing models to adapt to new tasks while updating a subset of parameters~\cite{anjum2024tensor}. Among these, Low-Rank Adaptation (LoRA)~\cite{hu2021lora} is particularly effective, reducing memory and computational overhead while maintaining fine-tuning efficiency (discussed in section~\ref{sec:lora}). Other techniques such as knowledge distillation~\cite{hinton2015distilling}, pruning~\cite{han2015deep}, and quantization~\cite{jacob2018quantization} are widely employed to compress models while preserving their core capabilities.

While pre-trained VLMs exhibit strong zero-shot capabilities, fine-tuning remains essential to align model representations with specific downstream tasks, such as image captioning, visual question answering, or multi-modal classification~\cite{lin2023vila}. Unlike pre-training, which requires massive, diverse datasets, fine-tuning typically involves smaller, domain-specific datasets, making it a more practical approach for adapting models to dynamic, real-world applications~\cite{sima2023drivelm}. This adaptability is particularly crucial for edge deployment scenarios, where VLMs must continuously learn from new data while operating under computational constraints.

\subsection{Federated Learning}
FL enables decentralized model fine-tuning by allowing clients to train models locally on distributed datasets, without sharing raw data with a central server. Each client performs local training and sends model updates—rather than data—to a central aggregator. Techniques such as FedAvg~\cite{mcmahan2017communication} and FedMekt~\cite{le2023fedmekt} are commonly used to merge these updates. This approach enhances data privacy, reduces infrastructure demands, and minimizes communication overhead by enabling multiple local training epochs. Additionally, FL improves model generalization across heterogeneous, non-iid client data distributions.

However, FL faces notable challenges. Limited compute resources on clients constrain model size and training complexity~\cite{yu2021toward}. Neuron drift, where activation patterns degrade due to the absence of certain classes during local updates, can harm model stability~\cite{wang2024fednlr}. Data heterogeneity across clients may further reduce global performance~\cite{agbaje2023fedcime}. Despite these hurdles, FL remains a promising strategy for scaling Vision-Language Models (VLMs) by enabling edge-based fine-tuning with built-in privacy preservation and adaptability.

We leverage FedAvg as the primary aggregation technique, weighting model updates by local dataset size to balance training efficiency and scalability. Given the resource constraints of edge devices, we focus on fine-tuning pre-trained models rather than training from scratch. Studies show that leveraging pre-trained models significantly reduce convergence time~\cite{tan2022federated} and resource consumption, so our FedVLM framework builds on a pre-trained foundation to accelerate fine-tuning in dynamic settings. 

Applying FL to VLMs introduces new challenges beyond those seen in unimodal FL for image or text classification. VLMs jointly process visual and textual inputs and often require optimization for both token classification and generation. While FL has been applied to multi-modal models~\cite{yu2023multimodal, zhao2022multimodal}, most prior efforts focus on classification tasks. Recent advances such as FedBiOT~\cite{wu2024fedbiot} demonstrate that using compressed adapters for large language model (LLM) fine-tuning can reduce resource demands on clients. However, fine-tuning multi-modal VLMs—distinct from unimodal LLMs—for reasoning and generative tasks remains underexplored.

Our proposed FedVLM framework addresses this gap by extending FL to multi-modal, generative VLMs. We develop scalable and privacy-preserving strategies for fine-tuning vision-language models on edge devices in dynamic environments.

\subsection{Personalized Federated Fine-Tuning with LoRA}
\label{sec:lora}
Low-rank Adaptation (LoRA), introduced by Hu et al.~\cite{hu2021lora}, enables efficient fine-tuning of large pre-trained models by injecting small trainable parameter matrices while keeping the original model weights frozen. Specifically, given a pre-trained weight matrix $W_0 \in \mathbb{R}^{m \times n}$, LoRA introduces two low-rank matrices: $A \in \mathbb{R}^{r \times n}$ and $B \in \mathbb{R}^{m \times r}$, where $r \ll \min(m, n)$. These low-rank matrices are optimized during training, allowing the model to adapt to new tasks with minimal computational cost and memory usage, significantly reducing the resources required for fine-tuning large models. By updating only these low-rank matrices, LoRA eliminates the need to modify the entire parameter set, making it highly effective for large-scale deployment in resource-constrained environments.

The initialization of these low-rank matrices plays a critical role in fine-tuning performance. Hu et al.~\cite{hu2021lora} demonstrated that initializing $B$ with zeros and $A$ with random Gaussian values yields more efficient training by approximating an identity transformation initially. This finding was further supported by Hayou et al.~\cite{hayou2024impact}, who demonstrated that this initialization outperforms alternatives such as initializing $A$ with zeros and $B$ randomly, reinforcing the impact of initialization strategies on LoRA's effectiveness.

Recent work have extended LoRA to FL to improve model personalization and efficiency across decentralized clients. Techniques such as FLoRA~\cite{nguyen2024flora} apply standard LoRA in FL, while HetLoRA~\cite{cho2024heterogeneous} allows each client to use distinct LoRA layers to address data heterogeneity. FDLoRA~\cite{qi2024fdlora} introduce dual low-rank layers to handle both homogeneous and heterogeneous data distributions, and FFA-LoRA~\cite{sun2024improving} simplifies the approach by aggregating only the $B$ matrix while freezing the $A$ matrix. Table~\ref{tab:comparison} summarizes the trade-offs among these approaches and highlights how our proposed approach, pLoRA, addresses current limitations.

\begin{table}[t]
\small
\centering
\caption{Comparison of LoRA-based methods in FL, analyzing communication overhead, dual-layer approaches, aggregation strategies, and support for pFL. The proposed pLoRA achieves low communication overhead while supporting pFL through selective matrix aggregation.}
\label{tab:comparison}
\scalebox{0.82}{
    \begin{tabular}{lcccc}
    \toprule
    \multirow{2}{*}{\textbf{Method}}
      & \textbf{Comm.}
      & \textbf{Dual-Layer} 
      & \textbf{Aggregating}
      & \multirow{2}{*}{\textbf{pFL}} \\   
      & \textbf{Overhead}
      & 
      & \textbf{Only $B$}
      &  \\
    \midrule
    FLoRA~\cite{nguyen2024flora}         
      & Moderate       
      & \xmark 
      & \xmark 
      & \xmark \\
    HetLoRA~\cite{cho2024heterogeneous}  
      & Moderate       
      & \xmark 
      & \xmark 
      & \cmark \\
    FDLoRA~\cite{qi2024fdlora}           
      & High  
      & \cmark 
      & \xmark 
      & \cmark \\
    FFA-LoRA~\cite{sun2024improving}     
      &  Low 
      & \xmark 
      & \cmark 
      & \xmark \\
    \midrule
    \textbf{pLoRA (Ours)}  
      &  Low 
      & \xmark 
      & \cmark 
      & \cmark \\
    \bottomrule
    \end{tabular}
}
\end{table}

Building on these advances, we propose pLoRA, which balances personalization and efficiency by strategically aggregating only the $B$ matrix—similar to FFA-LoRA—but retains client-specific $A$ matrices instead of freezing them. This hybrid strategy enhances client-level personalization while minimizing communication overhead, striking a balance between efficiency and flexibility. Unlike FFA-LoRA’s single-layer design, pLoRA enables finer-grained local adaptation, and compared to FDLoRA’s dual-layer setup, it reduces complexity without sacrificing performance.

pLoRA is well-suited for heterogeneous federated environments, where personalization is critical but communication and compute budgets are limited. Furthermore, our approach complements recent developments in personalized FL (pFL), such as knowledge-aware parameter coaching~\cite{zhi2024knowledge} and parameterized group knowledge transfer~\cite{zhang2021parameterized}. These methods offer promising directions for future extensions of our FedVLM framework, enabling adaptive VLM fine-tuning tailored to each client’s data distribution while preserving global knowledge consistency.

%% file: 3_methodology.tex
\section{Methodology}

\subsection{VLM Architecture} 
\label{sec:vlm}
A general VLM architecture consists of four key components: (1) a vision encoder to extract features from images, (2) a text encoder to process the associated questions or captions, (3) a projection layer to align the visual and textual embeddings, and (4) a language decoder to generate relevant responses.
Training VLMs typically follows a two stage process: pre-training and fine-tuning~\cite{zhou2024tinyllava}. The model learns to associate visual tokens with corresponding textual semantics in the pre-training phase, establishing foundational multimodal representations. The fine-tuning phase then refines these representations for specific downstream tasks, improving response accuracy and task-specific adaptation.
In this work, we focus exclusively on the fine-tuning phase within our FedVLM framework, which supports federated adaptation of VLM. Although we currently apply the framework only to fine-tuning, it can be extended to support both pre-training and fine-tuning in decentralized settings.

\vspace{2mm}
\noindent
\textbf{Florence2 as Base Model:} FedVLM builds upon Florence2~\cite{xiao2024florence}, a lightweight VLM with fewer than one billion parameters, making it well-suited for efficient fine-tuning in FL environments.
Unlike large-scale models such as BLIP-2~\cite{li2023blip} or GPT-4V~\cite{hurst2024gpt}, Florence2 offers a favorable trade-off between performance and resource efficiency. Pre-trained on high-quality annotated datasets, Florence2 is designed for tasks such as object detection, scene segmentation, and scene understanding. It uses a dual attention vision transformer (DaViT)~\cite{ding2022davit} as the vision encoder and BART~\cite{lewis2019bart} as the text encoder, following architectural designs of LLAVA and Tiny-LLAVA~\cite{zhou2024tinyllava}. Florence2’s compact size ensures lower communication cost and faster convergence during federated updates, which is critical in resource-constrained client environments.

\vspace{2mm}
\noindent
\textbf{Model Components and Fine-Tuning:}
The vision encoder, denoted as $V \in \mathbb{R}^{N_v \times D_v}$, processes images into visual embeddings, where $N_v$ represents the number of visual tokens and $D_v$ denotes the embedding dimension. The text encoder BART encodes the input text as a sequence of tokens, denoted as $T \in \mathbb{R}^{N_t \times D_t}$, where $N_t$ is the number of text tokens, and $D_t$ is the shared embedding dimension between the visual and text modalities. 

To align the visual and textual representations, a linear projection layer $P$ maps the visual embedding dimension $D_v$ to $D_t$. Empirical evidence suggests that this linear projection approach is more computationally efficient than additional transformer-based alignment layers~\cite{lin2023vila}, while maintaining competitive performance. The concatenated representation $\mathbf{X} = [P(V), T]$ integrates both modalities into a unified embedding space.
Finally, the language decoder generates the model’s response based on this aligned representation. The output token sequence is defined as $O = D_l(X)$, where $D_l$ represents the decoder function. Fine-tuning is guided by the cross-entropy loss $\mathcal{L}$, optimizing token prediction as follows:

\begin{align}
    \mathcal{L} = - \sum^{|o|}_{i = 1} log P_{\theta}(o_i | o_{<i}, x),
\end{align}
where $\theta$ represents the network parameters and $o$ represents the output tokens.

\subsection{FedVLM Fine-Tuning} 
Our proposed FedVLM framework (illustrated in Figure~\ref{fig:fedvlm}) enables federated fine-tuning of VLMs while minimizing communication and computational overhead. Each client begins with a pre-trained VLM that is either broadcasted by the server or acquired independently. We assume that clients have sufficient resources to fine-tune LoRA parameters for a few epochs and perform inference on local data. 
To reduce resource constraints, we freeze both encoder and decoder components, as these pre-trained modules already capture essential representations. 
Instead, we fine-tune only the pLoRA layer, a parameter-efficient module that modifies a subset of the model while maintaining overall stability. Model updates are aggregated at the central server using weighted averaging, following the update rule:

\begin{align}
    \mathcal{W}_{g}^{t+1} = \frac{1}{k} \sum_{k=1}^K \mathcal{W}_k^t 
\end{align}
where $\mathcal{W}_{g}^{t+1}$ is the updated global model after iteration $t+1$, $k$ is the total number of clients, and $\mathcal{W}_k^t$ denotes the model update from the client $k$ at iteration $t$. 

Our approach demonstrates that FL enables scalable VLM adaptation for real-world datasets. While this work focuses on fundamental FL fine-tuning, future optimizations—such as knowledge distillation, grouped query attention, shallow networks with increased depth~\cite{liu2024mobilellm}, and early fusion of text and images~\cite{team2024chameleon}, offer promising avenues for further efficiency improvements.

\textbf{Parameter-Efficient Fine-Tuning in FedVLM:} We leverage LoRA~\cite{hu2021lora}, a PEFT technique, to adapt pre-trained VLMs with minimal computational cost. LoRA keeps the majority of model parameters frozen while introducing trainable low-rank matrices $A$ and $B$ to model task-specific adaptations. Given a frozen pre-trained weight matrix $W_0$, the adapted transformation is:

\begin{align}
        h(x) &= W_0x + \Delta Wx = W_0x + BAx
\end{align}
where $h(x)$ is the hidden layer function, $\Delta W$ is the learned update, decomposed into matrices $B$ and $A$. LoRA starts with $B$ initialized to zero,  and $A$ is initialized with random Gaussian values. Since the product $BA$ is initially a zero-matrix, the fine-tuning process begins with the pre-trained model unchanged at initialization, allowing LoRA to incrementally adapt representations without catastrophic forgetting.
The decoder's output with LoRA is given by: 
\begin{align}
  O = D_{l:LoRA}(X)  
\end{align}
where $D_{l:LoRA}$ denotes the language decoder with LoRA-adapted parameters. This approach ensures that FedVLM fine-tunes large-scale VLMs efficiently in decentralized environments, preserving data privacy while enabling high-performance adaptation to client-specific tasks.

\subsection{pLoRA: Personalized LoRA}
To enhance local adaptability in heterogeneous (non-iid) data settings, we propose \textbf{pLoRA}, a personalized variant of LoRA that decomposes the update into shared and private components. pLoRA are integrated solely into the final layer of the VLM's language decoder, while keeping other pre-trained parameters frozen. This approach is designed to minimize computational overhead for each client, making it feasible for resource-constrained devices, such as edge devices, to perform fine-tuning with limited resources. By focusing adaptation on the final layer, we leverage the high-level representations learned by the pre-trained VLM, simplifying the personalization task for each client. This setup highlights the efficiency of pLoRA by reducing latency and resource usage, making fine-tuning scalable in federated environments with limited client resources.

The LoRA layers are optimized for two key purposes: \textbf{aggregation} and \textbf{personalization}. Each client locally fine-tunes its personalized LoRA weights, specifically the matrix $A$. These personalized weights are not shared during global aggregation; instead, only the aggregated LoRA matrix $B$ is communicated between clients via the server. The server updates the global model by combining the aggregated $B$ matrix with the original model weights $\mathcal{W}_o$ as follows:

\vspace{-0.5cm}
\begin{align} 
\mathcal{W}_{k}^{t+1} = \mathcal{W}_o + \mathcal{B}_g \mathcal{A}_p 
\end{align}
where $k$ is the client index, $\mathcal{W}_{k}^{t+1}$ denotes the updated weights of the personalized VLM, $\mathcal{B}_g$ is the globally aggregated LoRA $B$ matrix, and $\mathcal{A}_p$ is the personalized LoRA $A$ matrix, locally trained by clients. 

The aggregation of $\mathcal{B}_g$ matrix follows the FedAvg approach~\cite{mcmahan2017communication}, which is calculated as:

\vspace{-0.5cm}
\begin{align} 
\mathcal{B}_g^{t+1} = \frac{1}{k} \sum_{k=1}^K \mathcal{B}_k^t 
\end{align}

Thus, while the aggregated $B$ matrix is shared globally, each client retains its $A_p$ matrix. This allows for local fine-tuning based in specific data distributions while maintaining privacy during the global model update.

This design ensures efficient personalization while retaining global coherence, i.e., this selective sharing allows each client to personalize the model to its local data distribution while benefiting from shared global knowledge. Unlike prior work, such as FFA-LoRA and FLoRA, which either do not support personalization or share both $A$ and $B$, pLoRA uniquely personalizes $A$ while sharing only $B$. We empirically observe and justify this design: sharing $B$ stabilizes training across clients, while personalizing $A$ provides a sufficient client-specific capacity to adapt to non-iid data. Our design choice is supported both by empirical gains and the fact that $A$ captures the input-side projection, which is more sensitive to domain variation. Furthermore, while FFA-LoRA and FLoRA share architectural similarity, neither was designed for personalized federated VLM training, making our method distinct in both setting and objective.

\subsection{FedVLM Workflow}
Consider the workflow of a ($t+1$)-th round in FedVLM; which proceeds as follows:
\begin{enumerate}[label=\alph*)]
    \item \textbf{Local Training:} Each client starts training the locally available VLM using the personalized matrices $\mathcal{A}_p$, which remain local to the client in every round, and $\mathcal{B}$, which is received from the server after aggregated in round $t$. These pLoRA layers are trained on top of pre-trained weights from Florence2~\cite{xiao2024florence}. In the first round, $\mathcal{A}_p$ is randomly initialized, while $\mathcal{B}$ is initialized  as a zero matrix, ensuring that $\mathcal{B}\mathcal{A}_p = 0$ initially. 
    
    \item \textbf{Sending Updated:} After local training, the updated parameters of matrix $\mathcal{B}$ are sent to the server. 

    \item \textbf{Global Aggregation:} The server aggregates the received pLoRA layer parameters $\mathcal{B}$ from the clients using federated averaging.

    \item \textbf{Distribution:} The server distributes the aggregated parameters $\mathcal{B}_g^{t+1}$ to clients for use in next training round. 
\end{enumerate}

%% file: 4_experimental_setup.tex
\section{Experiments}

\subsection{Experimental Setup}

\textbf{Model}: We use Florence2 as our pre-trained model for fine-tuning with the following hyperparameters: a learning rate of $1\mathrm{e}{-6}$, a batch size of $16$, and $3$ local epochs per each client before aggregation.
The framework is implemented in ``PyTorch" with necessary libraries for pre-trained models retrieval and FL functionalities.

\vspace{2mm}
\noindent
\textbf{Dataset}: To evaluate the performance of FedVLM, we utilize a visual question-answering (VQA) task. Following the training methodology of tiny-LLaVA~\cite{zhou2024tinyllava}, we fine-tune the Florence2 model on reinforcement learning from AI feedback for vision (RLAIF-V) dataset~\cite{yu2024rlaif}. This dataset integrates high-quality AI-generated responses for image-based queries, combining samples from $14$ established datasets, including VQAv2~\cite{goyal2017making}, OK-VQA~\cite{marino2019ok}, and MSCOCO~\cite{lin2014microsoft}. Table~\ref{tab:dataset_counts} briefly summarizes the dataset distribution. 

To simulate data heterogeneity, each FL client is assigned data from a specific subset of the RLAIF-V dataset, mimicking real-world cases where organizations maintain domain-specific, siloed datasets. The dataset assignments follow a non-iid partitioning strategy, where clients receive images and queries primarily from distinct domains (e.g., medical or retail). This setup ensures rigorous evaluation in federated scenarios with extreme data heterogeneity.  

\begin{table}[t]
\centering
\caption{\textbf{RLAIF-V Dataset Distribution}: The dataset consists of samples from multiple standard datasets and is used to evaluate FedVLM in both IID and non-IID settings.}
\begin{tabular}{lr}
\toprule
\textbf{Original Dataset} & \textbf{Samples} \\
\midrule
LCS-558K & 15,956 \\
COCO & 15,199 \\
OK-VQA & 14,802 \\
VQAv2 & 12,942 \\
ART500K & 1,096 \\
\bottomrule
\end{tabular}
\label{tab:dataset_counts}
\end{table}

\vspace{2mm}
\noindent
\textbf{LoRA configuration}: We employ LoRA~\cite{hu2021lora} for parameter-efficient fine-tuning of Florence2’s language decoder. LoRA matrices are initialized using PyTorch defaults, with rank $r=4$ and a scaling factor of $8$. We apply a dropout rate of $0.1$ for stable training. In the federated setting, matrix $B$ is globally aggregated, while matrix $A$ is trained locally to enhance personalization. 

\vspace{2mm}
\noindent
\textbf{Hardware}: All experiments were conducted on an Ubuntu $18.04.6$ system equipped with an NVIDIA RTX A6000 GPU ($48$ GB VRAM), $52$ CPU cores, and $64$ GB RAM.

\subsection{Evaluation Strategies}

To assess FedVLM’s effectiveness, we investigate three key research questions: (i) \textbf{How does FedVLM compare to centralized training?}, (ii) \textbf{What is the impact of using pLoRA versus standard LoRA in non-iid settings?}, and (iii) \textbf{How does the framework scale as the number of clients increases?} Below, we summarize our findings.

\begin{figure}[bh]
    \begin{subfigure}[b]{0.235\textwidth}
       \centering
        \includegraphics[height=0.7\columnwidth, width=\columnwidth]{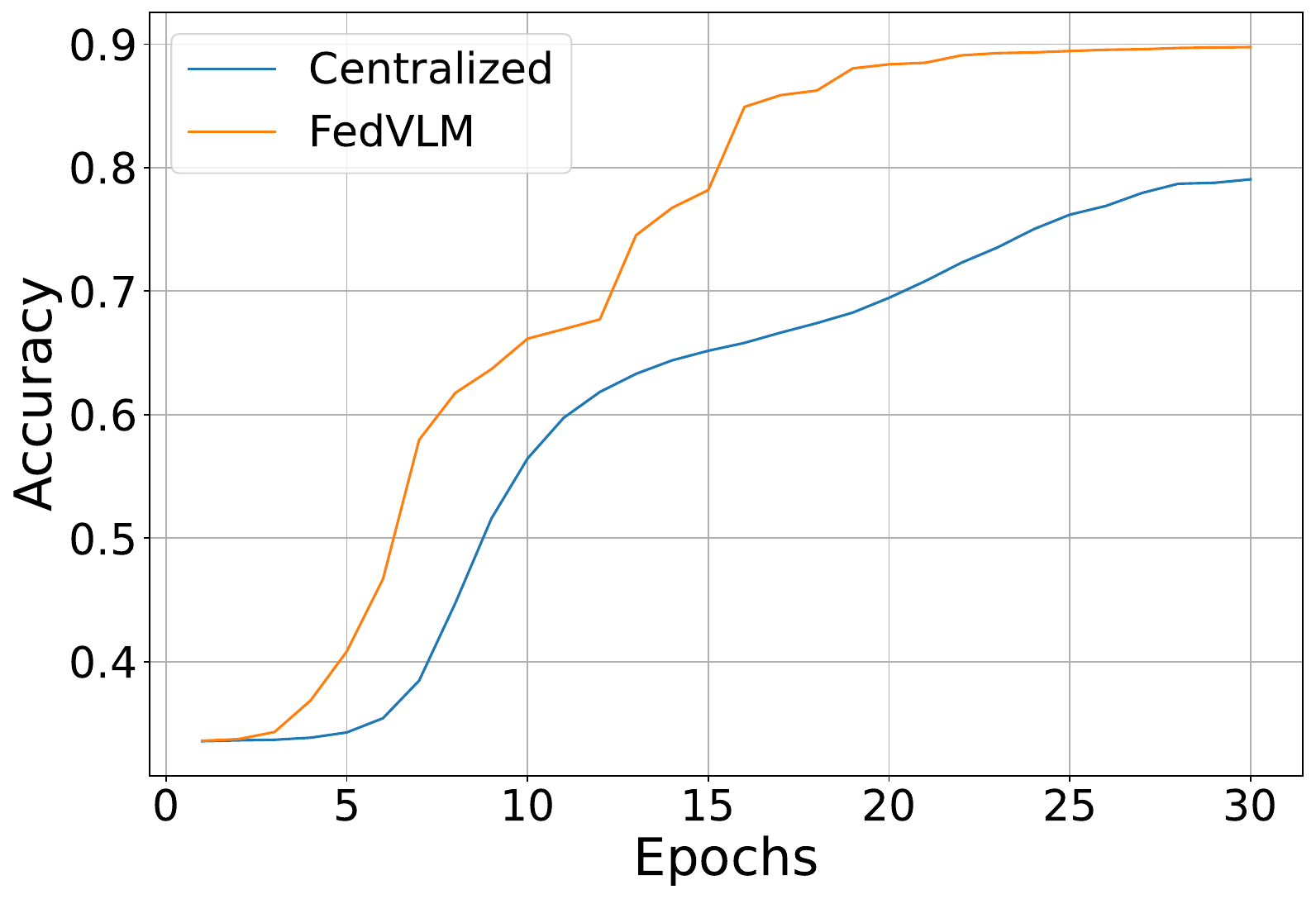}
        \caption{OK-VQA}
        \label{fig:centvsfed}
    \end{subfigure}
    \begin{subfigure}[b]{0.235\textwidth}
        \centering
        \includegraphics[height=0.7\columnwidth, width=\columnwidth]{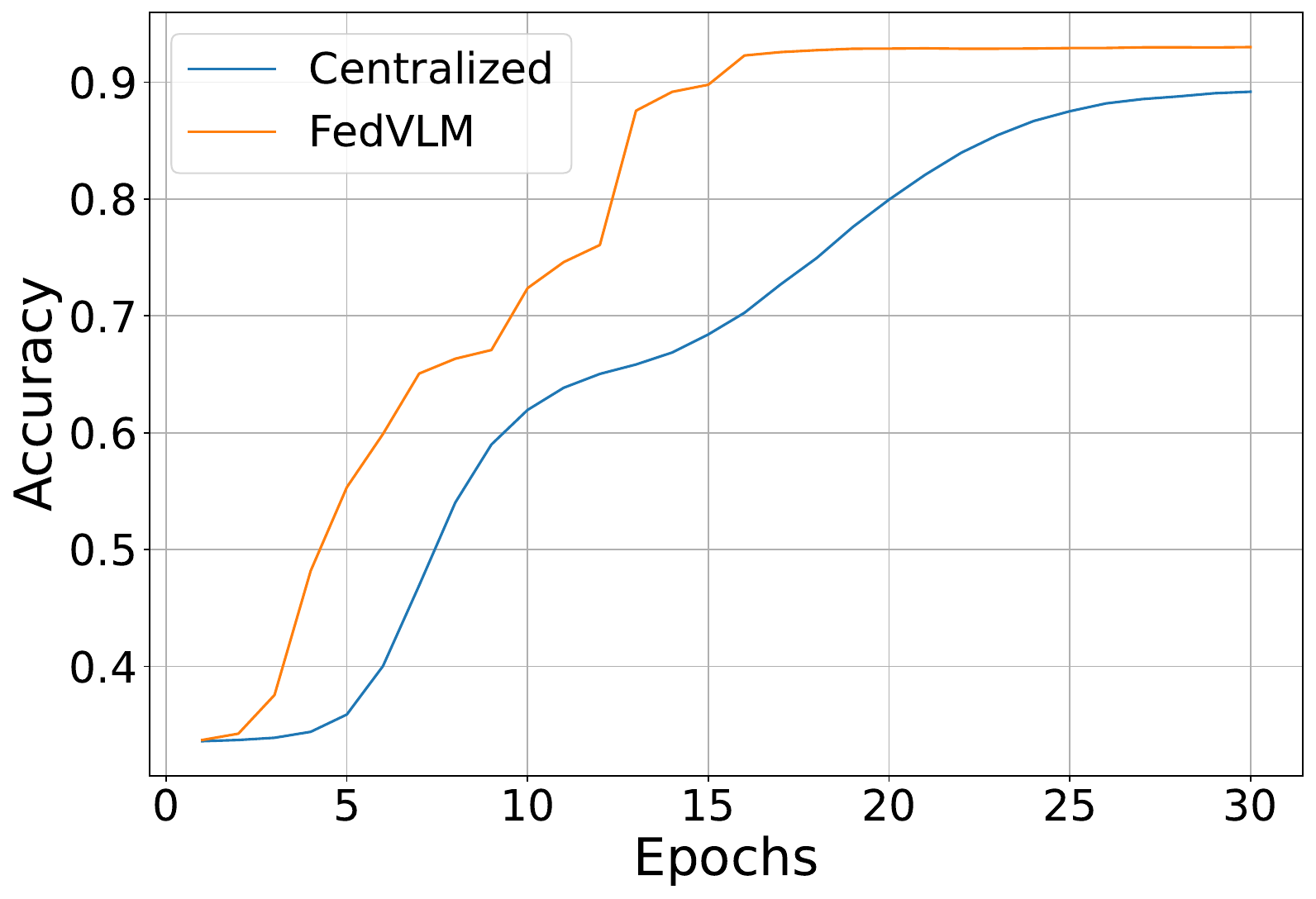}
        \caption{RLAIF-V}
        \label{fig:centvsfed_rlaifv}
    \end{subfigure}
    \vspace{3mm}
    \caption{\textbf{Federated vs. Centralized Performance Analysis:} We compare the convergence rates and accuracy of FedVLM and centralized training. FedVLM demonstrates faster convergence and higher accuracy, illustrating its effectiveness in FL environments.}
    \label{fig:centanalysis}
    \vspace{3mm}
\end{figure}

\vspace{2mm}
\noindent
\textbf{Centralized vs. FedVLM}: We compare FedVLM with central training using LoRA on four clients (three local epochs per round). This experiment demonstrates that FL can achieve comparable or superior results without requiring a large centralized infrastructure. On the OK-VQA dataset~\cite{marino2019ok} (iid setting), FedVLM converges faster and improves accuracy (see figure~\ref{fig:centvsfed}). Similarly, on the full RLAIF-V dataset, centralized training reaches 89\% accuracy, whereas FedVLM not only converges more rapidly but also delivers superior overall performance (see figure~\ref{fig:centvsfed_rlaifv}).

\begin{table}[t]
    \centering
        \caption{\textbf{Accuracy comparison of pLoRA against SOTA:} Our findings illustrate that pLoRA improves the average performance for every client in non-IID and IID settings.}
    \begin{tabular}{lccc}
        \toprule
         \textbf{Data distribution}&\textbf{pLoRA}& \textbf{FLoRA}  & \textbf{FFA-LoRA}\\
         \midrule
         Non-IID&  0.867&  0.696 & 0.343\\
         IID&  0.745&  0.647 & 0.337\\ 
         \bottomrule
    \end{tabular}
    \label{tab:ploraiidniid}
\end{table}

\begin{figure}[t]
    \begin{subfigure}[b]{0.235\textwidth}
       \centering
        \includegraphics[height=0.7\columnwidth, width=\columnwidth]{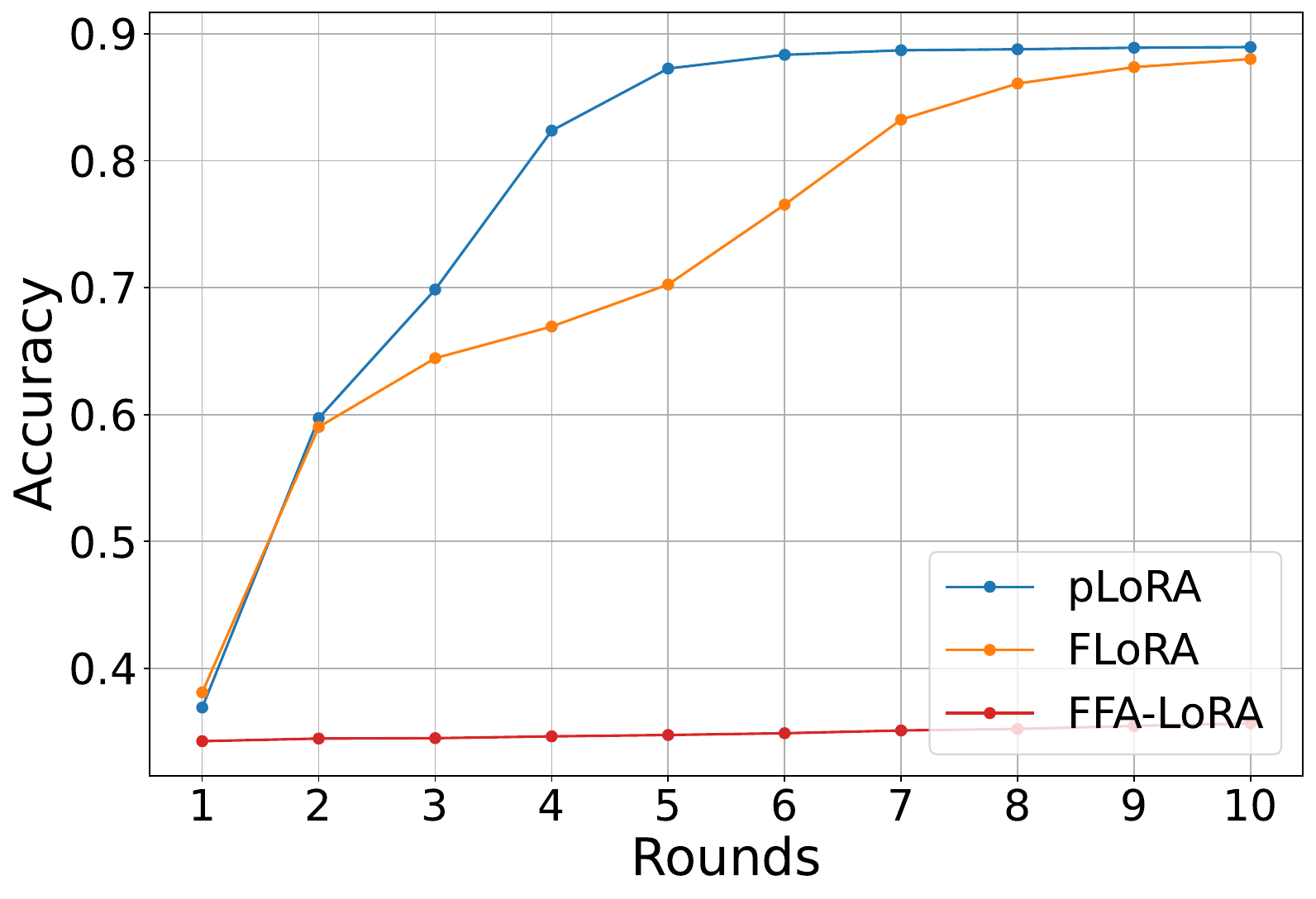}
        \caption{ Accuracy comparison }
        \label{fig:plorasotaacc}
    \end{subfigure}
    \begin{subfigure}[b]{0.235\textwidth}
        \centering
        \includegraphics[height=0.7\columnwidth, width=\columnwidth]{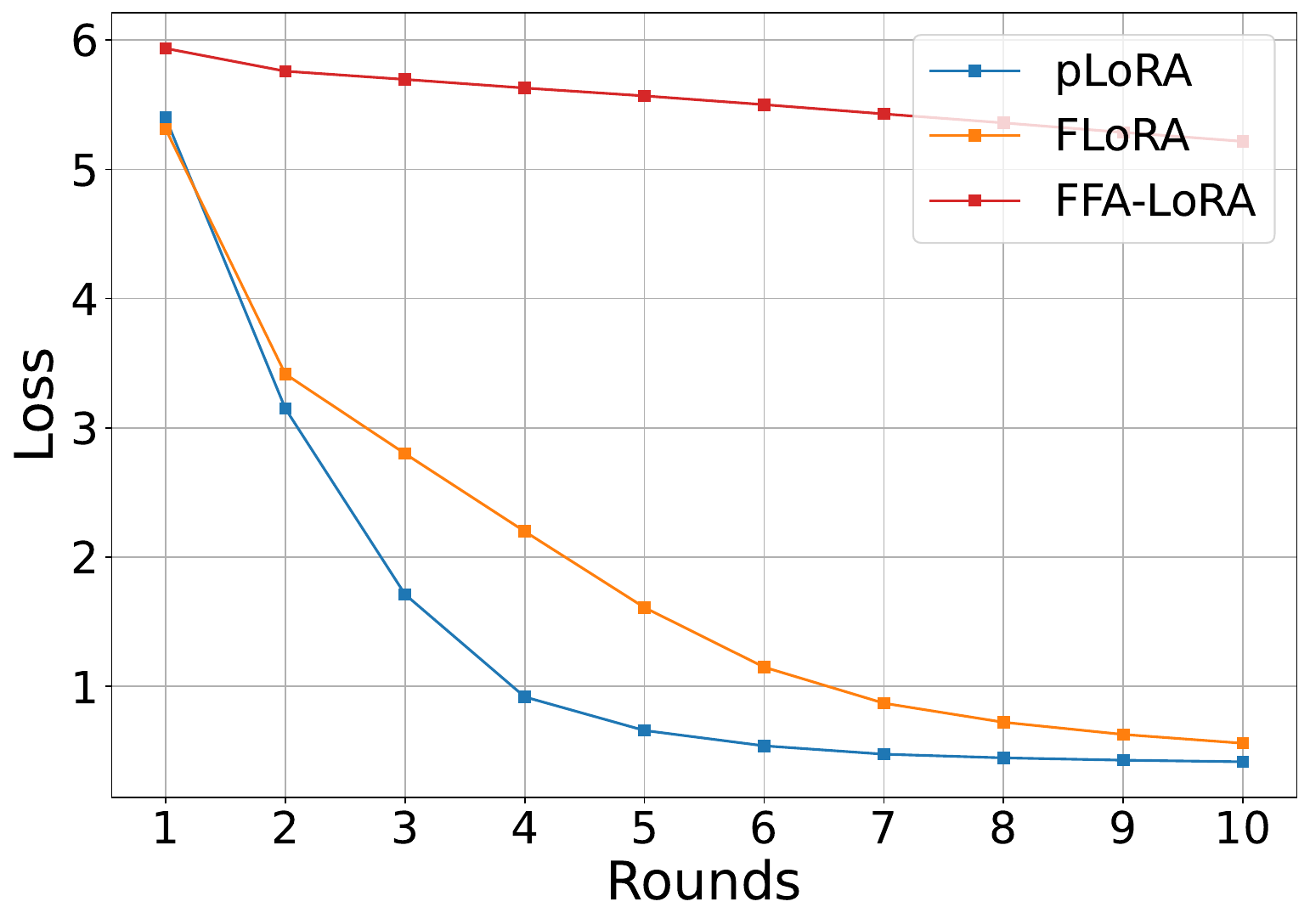}
        \caption{Loss comparison}
        \label{fig:plorasotaloss}
    \end{subfigure}
    \vspace{3mm}
    \caption{\textbf{Performance Analysis Against SOTA}: pLoRA demonstrates substantial performance gains over both standard LoRA and FFA-LoRA, underscoring its effectiveness in FL settings.}
    \label{fig:hybrid}
    \vspace{6mm}
\end{figure}

\begin{figure}[h]
    \centering
    \includegraphics[height=0.5\columnwidth, width=\columnwidth]{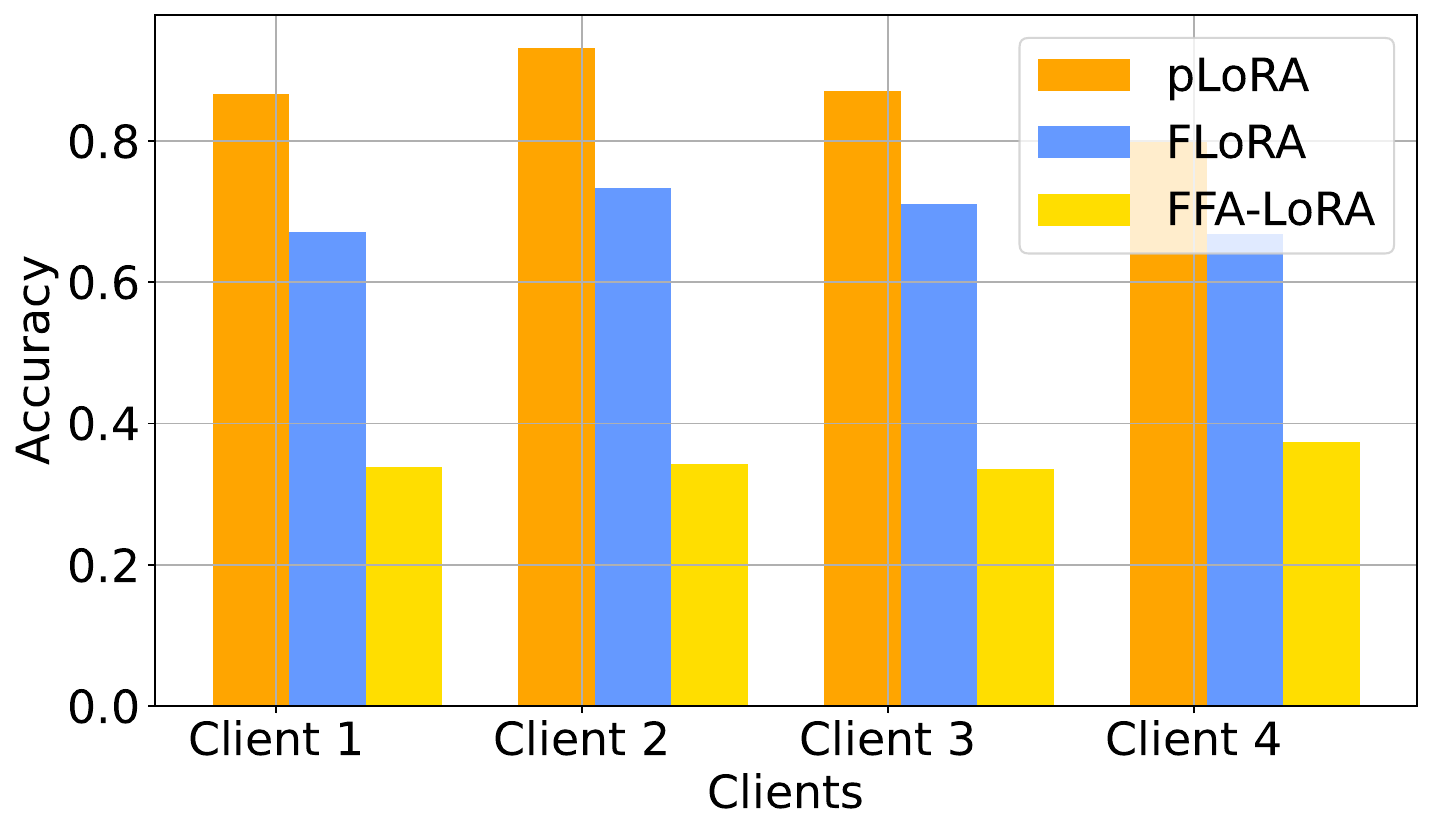}
    \caption{\textbf{Performance Comparison Across Client:} We show pLoRA's improvement over SOTA methods for each client in non-IID settings, demonstrating consistent benefits in personalized FL.}
    \label{fig:noniidbar}
    \vspace{6mm}
\end{figure}

\vspace{2mm}
\noindent
\textbf{pLoRA vs. SOTA}: We evaluate our proposed pLoRA against standard LoRA~\cite{hu2021lora} and FFA-LoRA~\cite{sun2024improving} under identical training conditions.
Since FLoRA~\cite{nguyen2024flora} was the first to integrate standard LoRA into FL, our comparison with standard LoRA also benchmarks against FLoRA’s approach. 
Experiments were conducted in both iid and non-iid settings with four clients, each performing three local epochs across five rounds. As summarized in table~\ref{tab:ploraiidniid}, our results show that pLoRA within FedVLM achieves: (i) 24.5\% higher accuracy in non-iid settings and (ii) 15.1\% higher accuracy in iid settingscompared to existing methods. While current SOTA methods eventually reach similar accuracy, they require significantly more communication rounds to achieve the same performance. Further insights on loss reduction and accuracy improvements are presented in figure~\ref{fig:hybrid}, while individual client performance in non-iid settings is shown in figure~\ref{fig:noniidbar}.

For FFA-LoRA, we follow the authors’ recommended configuration, using a rank and scaling factor of $8$. Notably, FFA-LoRA exhibits slower convergence on the RLAIF-V dataset, likely due to the uniformly applied small learning rate ($1\mathrm{e}{-6}$), ensuring a fair comparison. Increasing the learning rate improves convergence speed but results in unstable optimizations and undesired local minima. These findings indicate that pLoRA enhances model personalization in non-iid FL while maintaining minimal training parameters, ultimately improving VLM performance for client-specific tasks.

\begin{figure*}[t]
    \centering
    \begin{subfigure}[b]{0.245\linewidth}
        \centering
        \includegraphics[height=0.7\columnwidth, width=\textwidth]{non_iid_2_client_acc_val.pdf}
        \caption{2 clients}
        \label{fig:niidacc2}
    \end{subfigure}
    \hfill
    \begin{subfigure}[b]{0.245\linewidth}
        \centering
        \includegraphics[height=0.7\columnwidth, width=\textwidth]{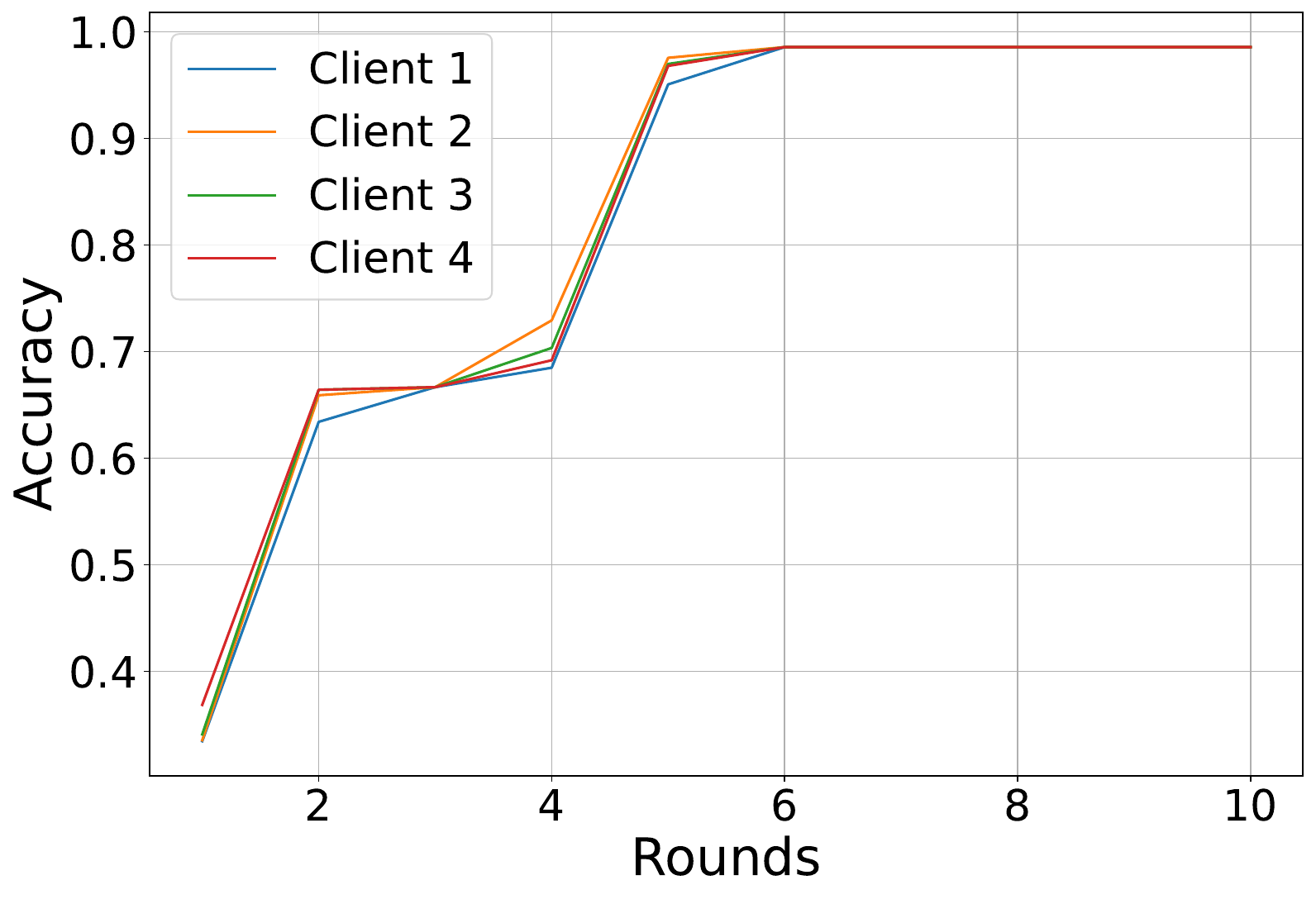}
        \caption{4 clients}
        \label{fig:niidacc4}
    \end{subfigure}
    \hfill
    \begin{subfigure}[b]{0.245\linewidth}
        \centering
        \includegraphics[height=0.7\columnwidth, width=\textwidth]{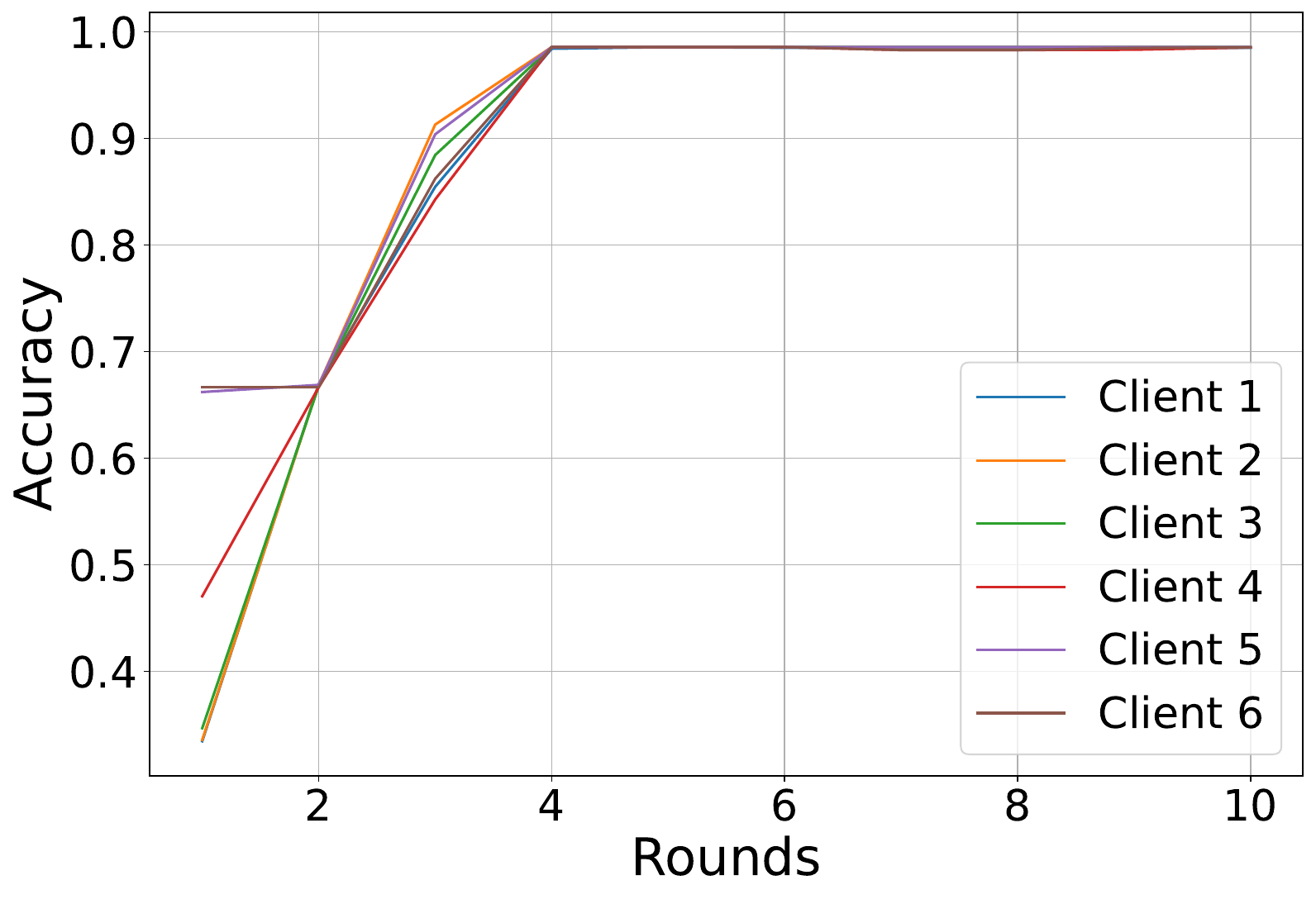}
        \caption{6 clients}
        \label{fig:niidacc6}
    \end{subfigure}
    \hfill
    \begin{subfigure}[b]{0.245\linewidth}
        \centering
        \includegraphics[height=0.7\columnwidth, width=\textwidth]{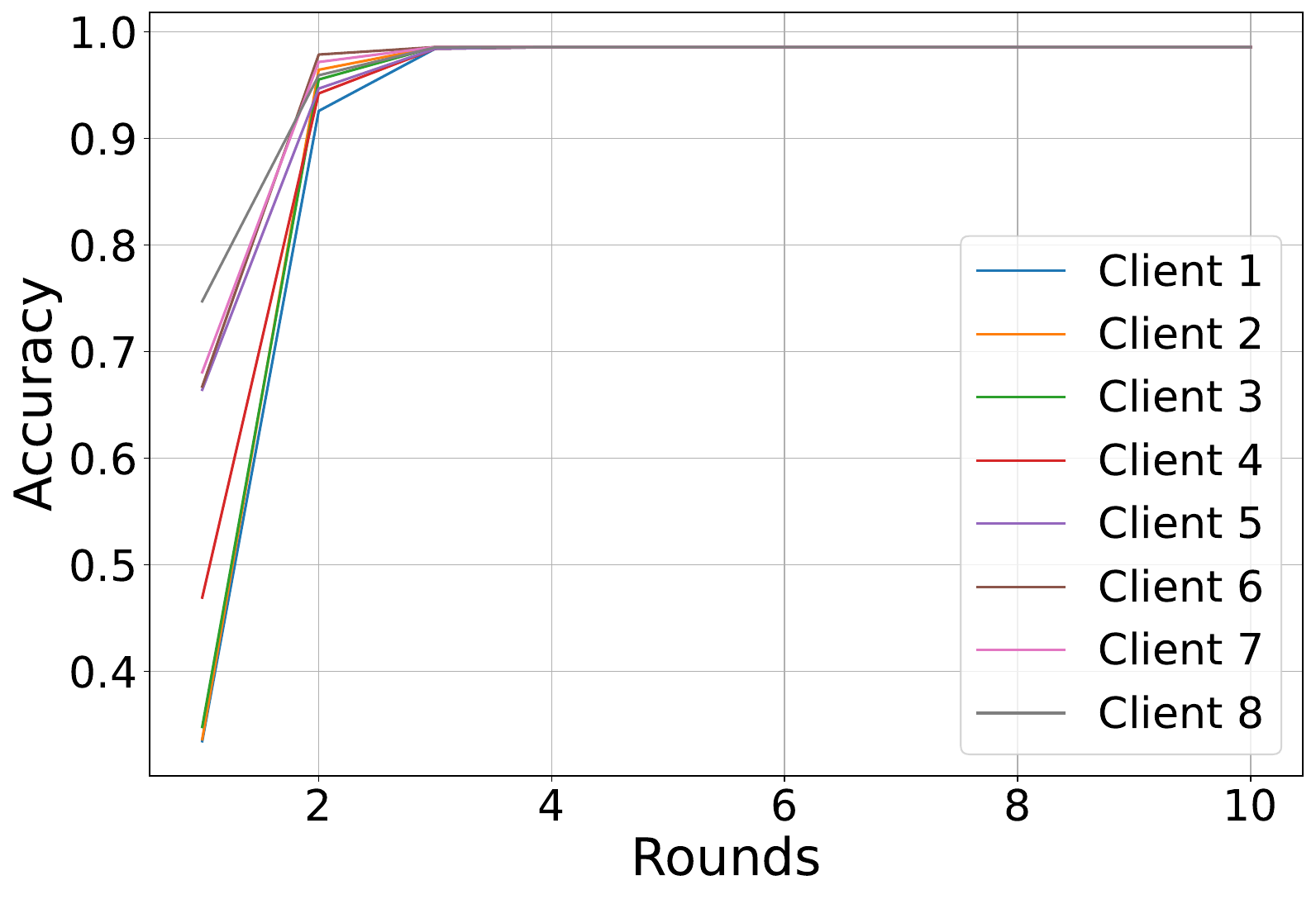}
        \caption{8 clients}
        \label{fig:niidacc8}
    \end{subfigure}
    \vspace{2mm}
    \caption{\textbf{Accuracy Performance Across Different Client Sizes:} The accuracy curves illustrate FedVLM’s improvements with pLoRA and federated learning, highlighting consistent accuracy gains across varying numbers of clients.}
    \label{fig:niid_acc}
    
\end{figure*}

\begin{figure*}[h!]
    \centering
    \begin{subfigure}[b]{0.245\linewidth}
        \centering
        \includegraphics[height=0.7\columnwidth, width=\textwidth]{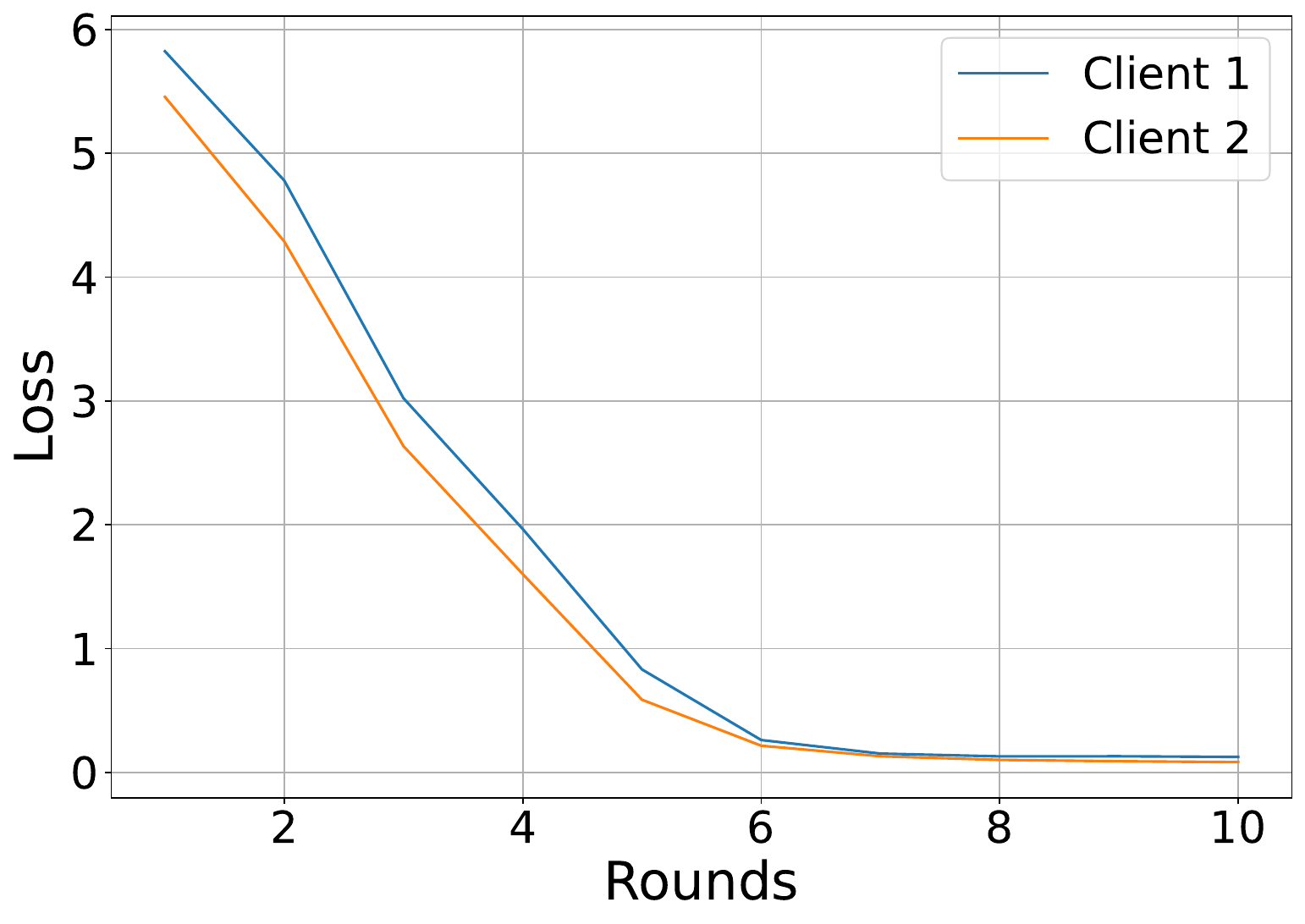}
        \caption{2 clients}
        \label{fig:niidloss2}
    \end{subfigure}
    \hfill
    \begin{subfigure}[b]{0.245\linewidth}
        \centering
        \includegraphics[height=0.7\columnwidth, width=\textwidth]{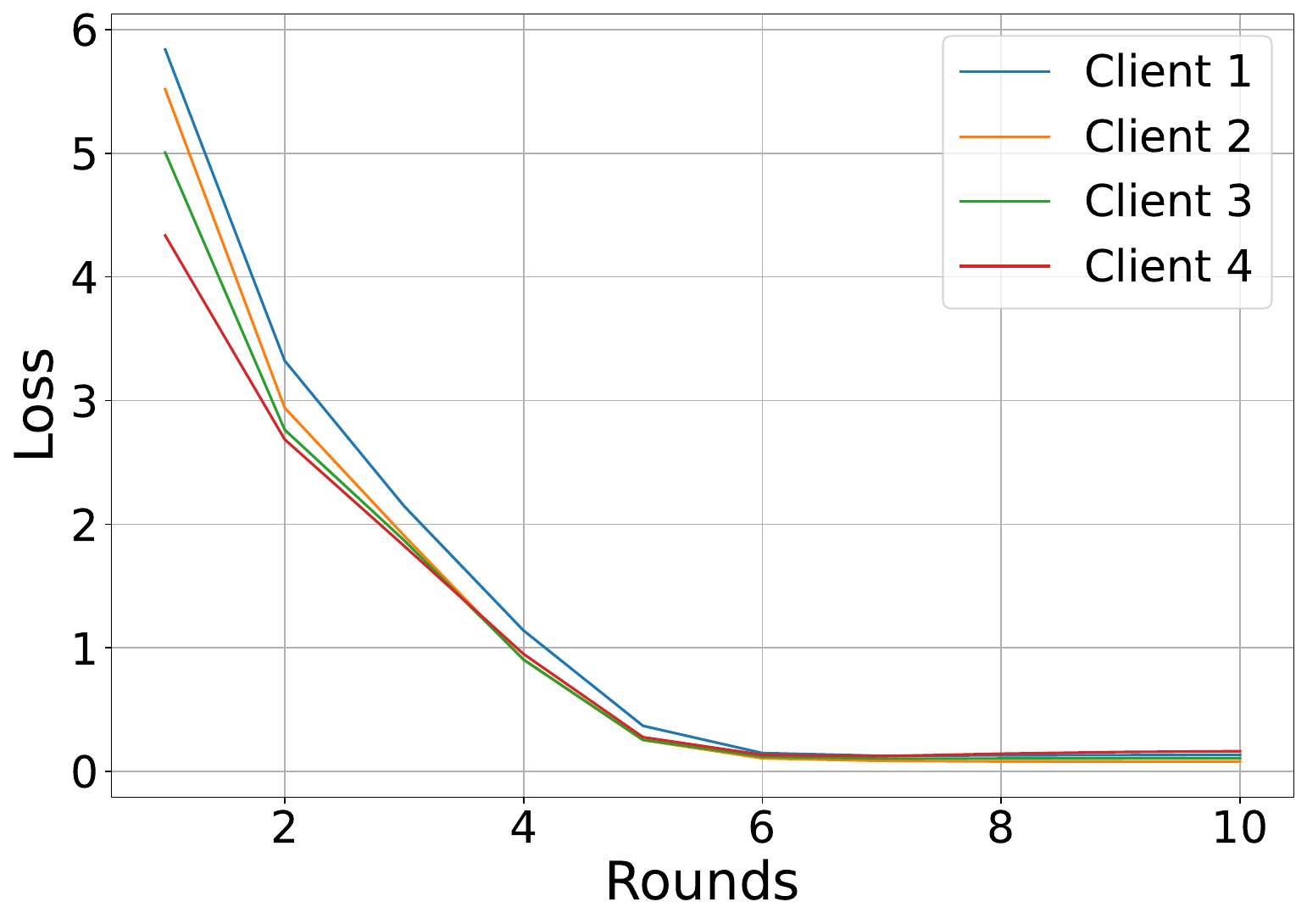}
        \caption{4 clients}
        \label{fig:niidloss4}
    \end{subfigure}
    \hfill
    \begin{subfigure}[b]{0.245\linewidth}
        \centering
        \includegraphics[height=0.7\columnwidth, width=\textwidth]{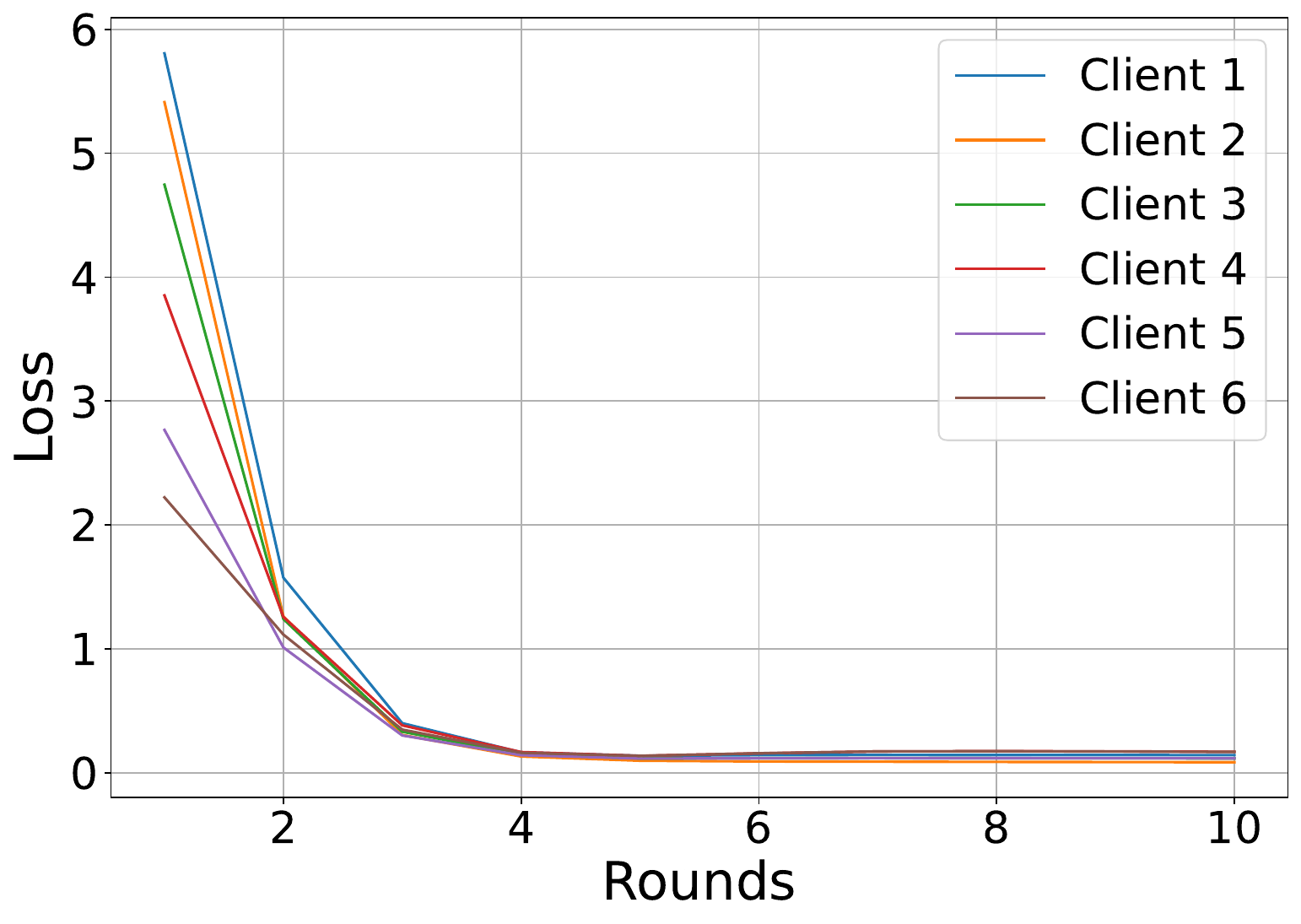}
        \caption{6 clients}
        \label{fig:niidloss6}
    \end{subfigure}
    \hfill
    \begin{subfigure}[b]{0.245\linewidth}
        \centering
        \includegraphics[height=0.7\columnwidth, width=\textwidth]{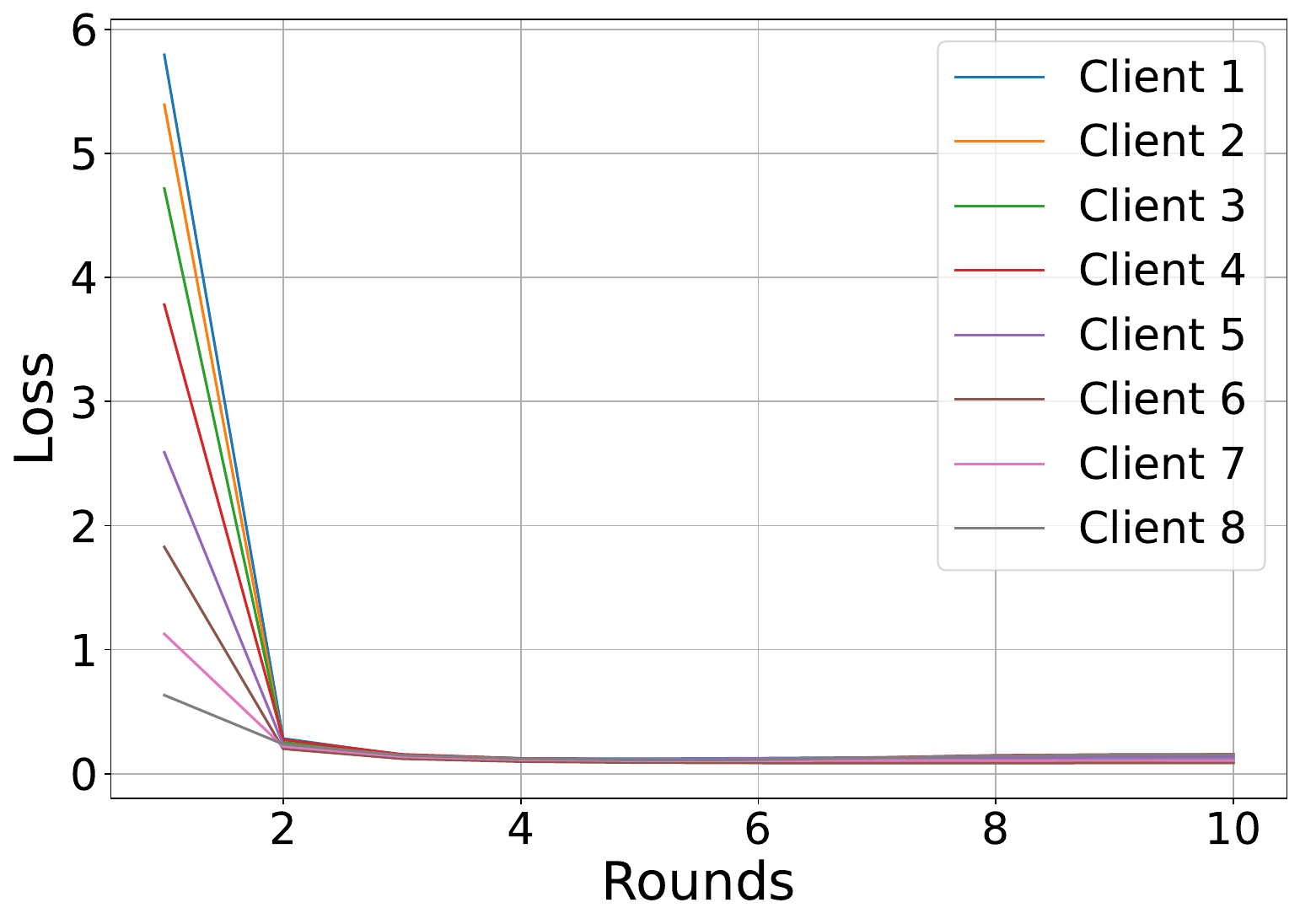}
        \caption{8 clients}
        \label{fig:niidloss8}
    \end{subfigure}
    \vspace{2mm}
    \caption{\textbf{Loss Curves Across Different Client Sizes:} The loss curves show smooth convergence, reflecting FedVLM's performance improvements and stability across different client configurations.}
    \label{fig:niid_loss}
    \vspace{2mm}
\end{figure*}

\vspace{2mm}
\noindent
\textbf{Client ablation}: We evaluate FedVLM’s performance across varying numbers of clients (2, 4, 6, and 8), each trained on different sub-datasets to simulate non-iid conditions. Results, summarized in table~\ref{tab:abclient2} using standard evaluation metrics (accuracy, recall, precision, and F1-score), demonstrate that FedVLM maintains robust performance across different client counts. Additionally, we observe that personalized VLMs in FedVLM converge faster as the number of clients increases, as depicted in the figures~\ref{fig:niidacc2},~\ref{fig:niidacc4},~\ref{fig:niidacc6}, and~\ref{fig:niidacc8}. The corresponding loss curves for each client configuration are presented in the figures~\ref{fig:niidloss2},~\ref{fig:niidloss4},~\ref{fig:niidloss6}, and~\ref{fig:niidloss8}.

\begin{figure}[h]
    \begin{subfigure}[b]{0.235\textwidth}
       \centering
        \includegraphics[height=0.7\columnwidth, width=\columnwidth]{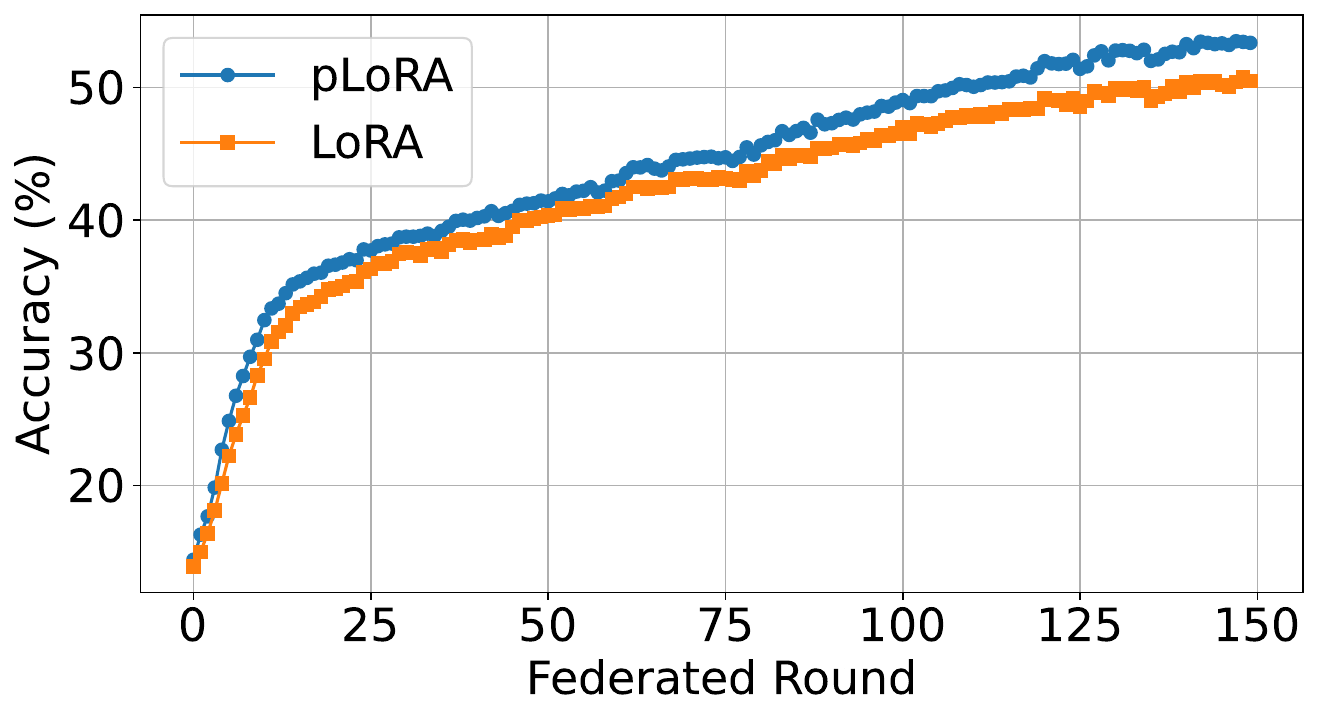}
        \caption{ Accuracy comparison }
        \label{fig:ploraloraacc}
    \end{subfigure}
    \begin{subfigure}[b]{0.235\textwidth}
        \centering
        \includegraphics[height=0.7\columnwidth, width=\columnwidth]{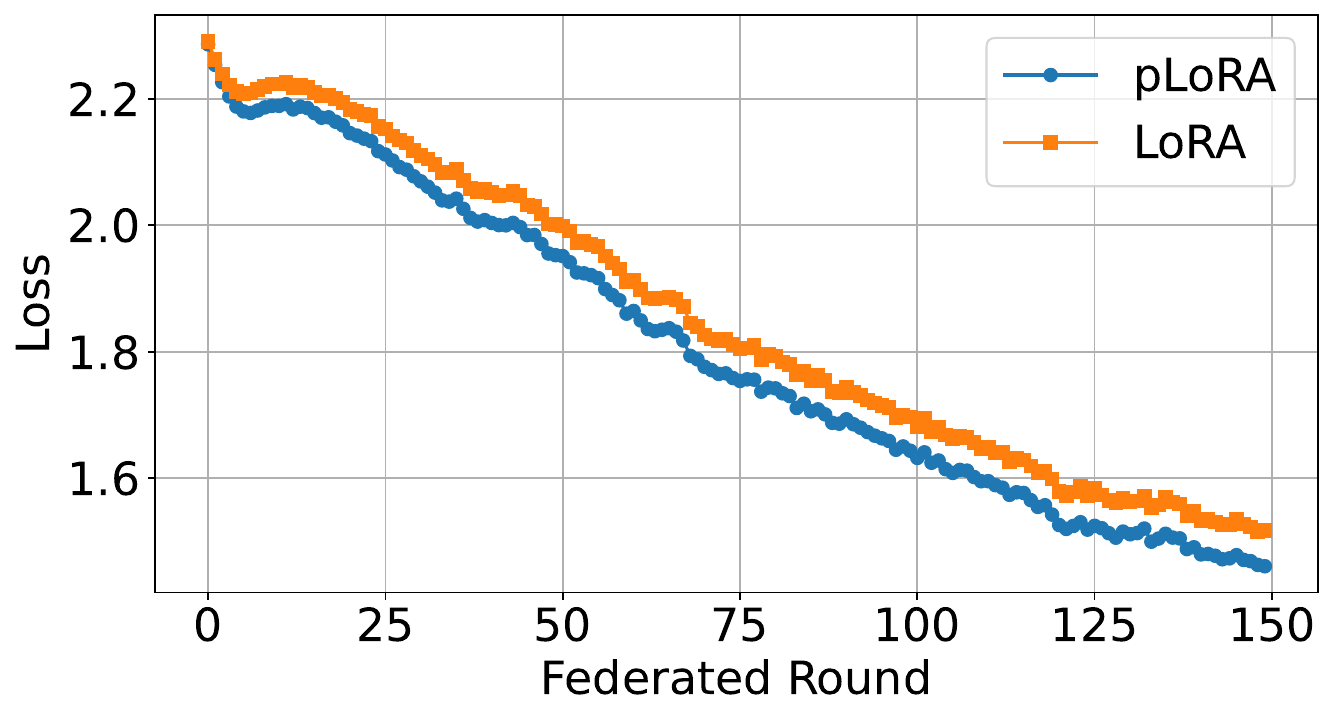}
        \caption{Loss comparison}
        \label{fig:ploraloraloss}
    \end{subfigure}
    \vspace{2mm}
    \caption{\textbf{Performance Analysis Against LoRA}: pLoRA demonstrates performance gains over standard LoRA in CIFAR-10, underscoring its effectiveness in FL settings.}
    \label{fig:plora_vs_lora}
    \vspace{4mm}
\end{figure}

\begin{table}[t]
    \centering
    \caption{\textbf{Client Ablation Study in Non-IID Setting:} Each client is assigned a distinct dataset to simulate data heterogeneity. We compare model performance before and after fine-tuning (ft), highlighting personalization impact. Results show that FedVLM maintains consistent performance across varying client counts.}
    \scalebox{0.85}{ 
        \begin{tabular}{cccccc}
            \toprule
            \textbf{Number} & \textbf{Accuracy} & \textbf{Precision} & \textbf{Recall} & \textbf{F1-Score} & \textbf{Loss} \\
            \textbf{of Clients} & \textbf{(Avg.)} & \textbf{(Avg.)} & \textbf{(Avg.)} & \textbf{(Avg.)} & \textbf{(Avg.)} \\
            \midrule
            2 w/o ft& 0.33& 0.66& 0.33& 0.34& 5.92\\
            2 & 0.98 & 0.97 & 0.98 & 0.97 &0.11 \\
             \midrule
            4 w/o ft& 0.33& 0.66& 0.33& 0.34&5.92\\
            4 & 0.98 & 0.97 & 0.98 & 0.97 & 0.13 \\
             \midrule
            6 w/o ft& 0.33& 0.66& 0.33& 0.34&5.92\\
            6 & 0.98 & 0.97 & 0.98 & 0.97 & 0.13 \\
             \midrule
            8 w/o ft& 0.33& 0.66& 0.33& 0.34&5.92\\
            8 & 0.98 & 0.97 & 0.98 & 0.97 & 0.12 \\
            \bottomrule
        \end{tabular}
    }
    \label{tab:abclient2}
\end{table}

\vspace{2mm}
\noindent
\textbf{Rank Ablation}: To assess the impact of rank selection in pLoRA, we evaluate different rank settings ($r\in[2,4,8,16]$) in a non-iid scenario with two clients over 10 rounds. Since increasing rank can introduce latency overhead while preserving critical information, we follow prior work~\cite{hu2021lora, sun2024improving} to determine an optimal balance. As shown in table~\ref{tab:abrank}, increasing the rank has minimal impact on performance, likely due to fine-tuning occurring at the final layer. However, these results reaffirm pLoRA’s effectiveness in improving model performance while maintaining computational efficiency. 

Furthermore, we validate FedVLM’s scalability by comparing pLoRA with LoRA under non-iid FL settings. Using a MobileNetV3~\cite{howard2019searching} model pre-trained on ImageNet-1K, we fine-tune on CIFAR-10~\cite{krizhevsky2009learning}. As shown in figure~\ref{fig:plora_vs_lora}, pLoRA consistently outperforms LoRA across all communication rounds due to its enhanced personalization capabilities. The experiment involves 40 clients trained over 150 rounds (each with three local epochs), with CIFAR-10 partitioned into 80 shards (two per client), randomly assigned to simulate non-iid conditions.  

\begin{table}[t]
    \centering
    \caption{\textbf{Rank Ablation Study in Non-IID Setting:}  Results suggest that varying the matrix rank has minimal impact on FedVLM's performance on the RLHAIF-V dataset, reinforcing pLoRA’s adaptability to different rank configurations.}
    \begin{tabular}{lcccc}
        \toprule
         \textbf{Rank}&  \textbf{2}&  \textbf{4} & \textbf{8}&\textbf{16}\\
         \midrule
         Client  1 (Acc.)&  0.985&  0.9857& 0.985&0.985\\ 
         Client 2 (Acc.)&  0.9857&  0.9857& 0.985&0.9857\\
         \hline
         Client 1 (Loss)&  0.1175&  0.1176& 0.117&0.116\\ 
         Client 2 (Loss)&  0.0754&  0.075& 0.0759&0.0758\\ 
         \bottomrule
    \end{tabular}
    \label{tab:abrank}
\end{table}

\begin{figure}
    \begin{subfigure}[b]{0.235\textwidth}
        \centering
        \includegraphics[height=0.7\columnwidth, width=\columnwidth]{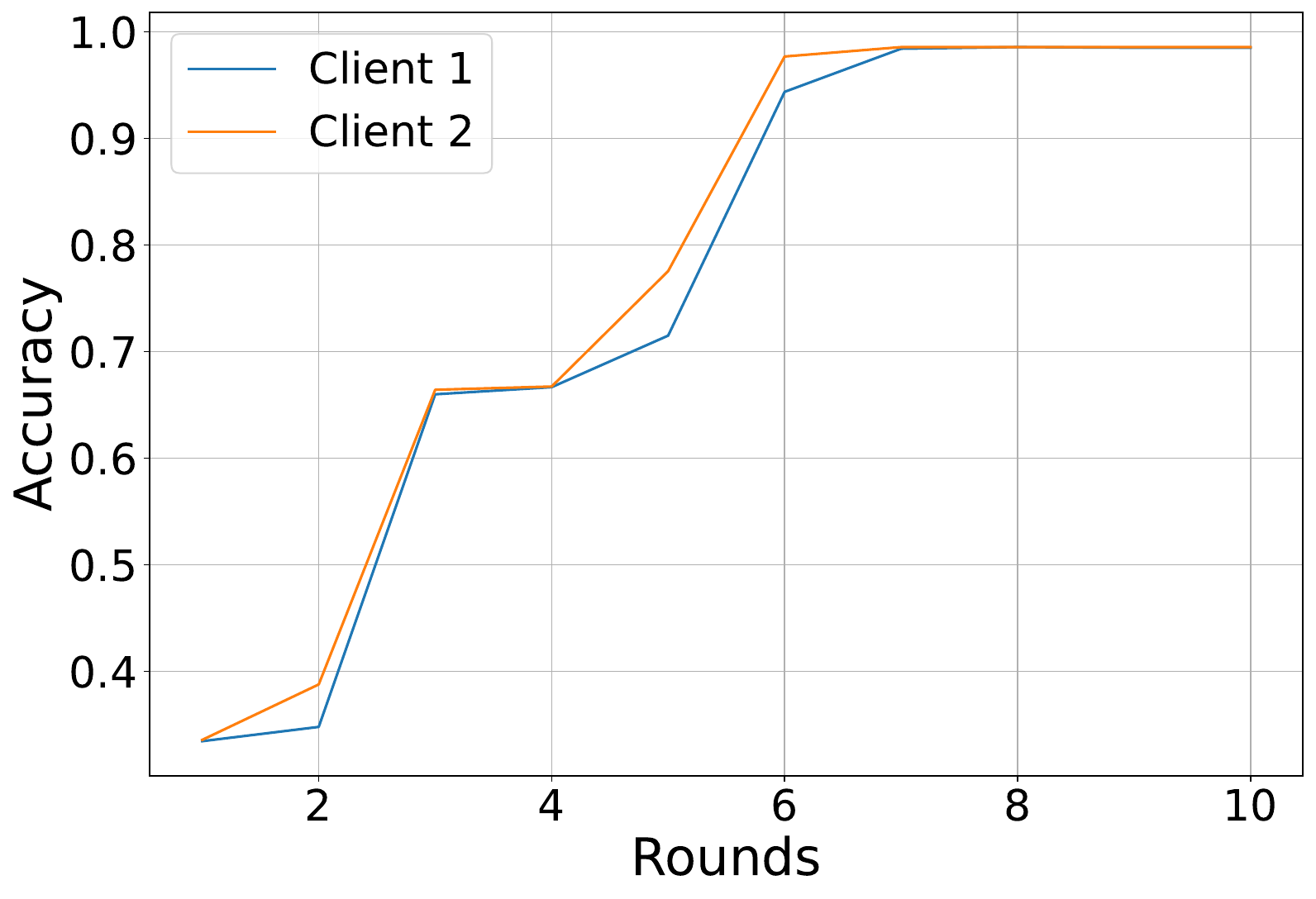}
        \caption{FedAvg}
        \label{fig:fedavgacc}
    \end{subfigure}
    \begin{subfigure}[b]{0.235\textwidth}
        \centering
        \includegraphics[height=0.7\columnwidth, width=\columnwidth]{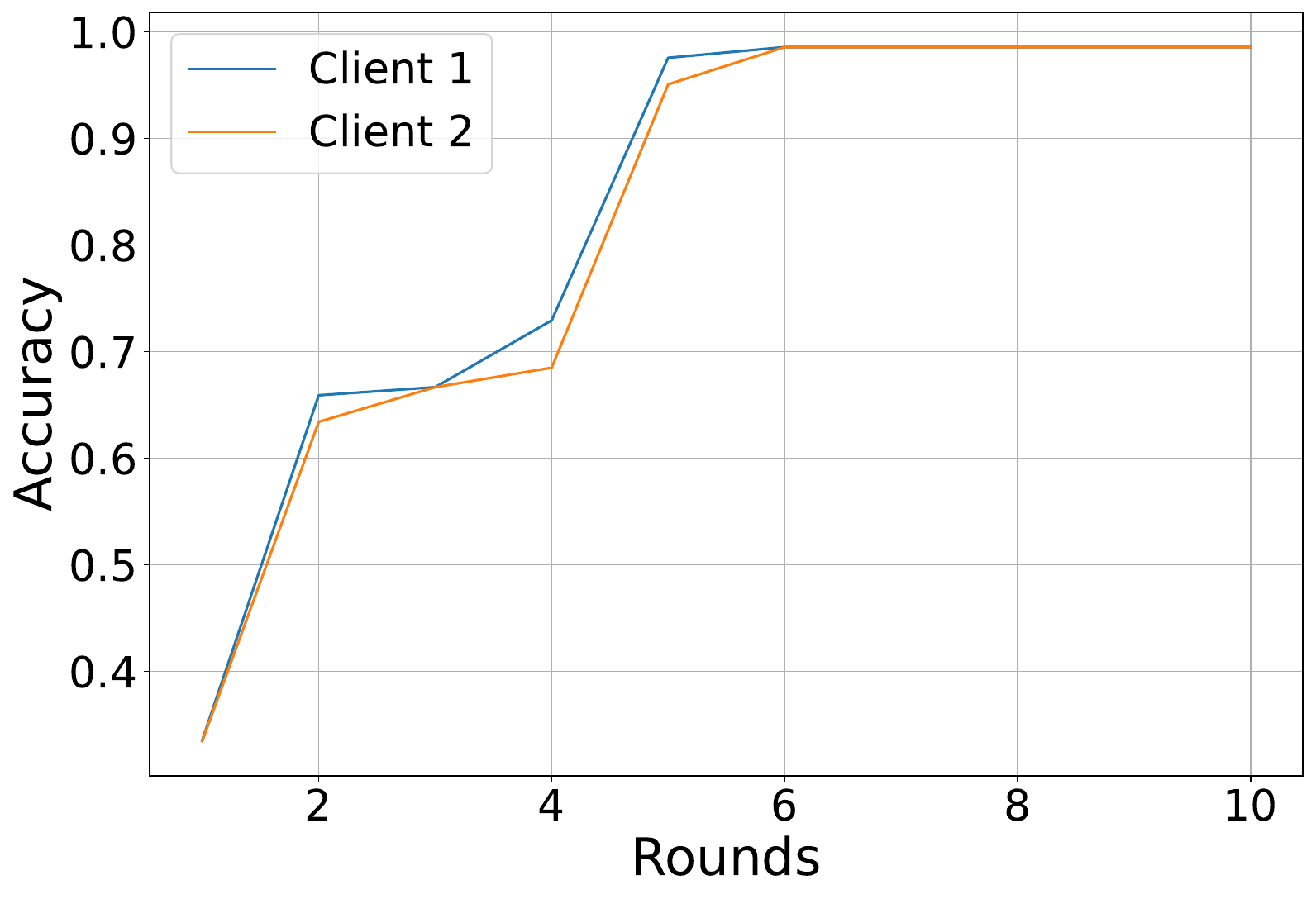}
        \caption{FedProx}
        \label{fig:fedproxacc}
    \end{subfigure}
    \vspace{2mm}
    \caption{\textbf{Comparison of Aggregation Methods:} FedProx mitigates data distribution shifts among clients by incorporating a proximal term in local updates, enhancing model stability in federated settings.}
    \label{fig:aggab}
    \vspace{2mm}
\end{figure}

\vspace{2mm}
\noindent
\textbf{Aggregation}: To demonstrate flexibility in aggregation strategies, we compare FedVLM with FedProx~\cite{li2020federated}, a widely used method that mitigates distribution shifts through a proximal term that constrains local updates, reducing model drift in heterogeneous federated settings. 
Our results confirm that FedProx effectively limits model drift while maintaining a balance between personalization and global model performance. This further highlights FedVLM’s adaptability to environments with high data variability across clients, illustrated in figure~\ref{fig:aggab}.

\subsection{Analysis}
This section outlines the motivations for employing FL to enhance and personalize VLMs. We highlight FL’s advantages over centralized training and justify the use of personalized LoRA adapters in our approach. 

\vspace{2mm}
\noindent
\textbf{Why FedVLM for Improving VLMs?}: Centralized VLM training demands extensive computational resources and costly infrastructure, often inaccessible to small and mid-sized organizations. These barriers restrict broader access to SOTA VLMs. In contrast, FL enables distributed model training across decentralized devices, reducing reliance on centralized hardware while preserving data privacy.  

Our proposed FedVLM framework leverages this distributed approach, enabling organizations to train models locally. democratizes access to advanced VLMs, fostering the development of efficient, on-device personalized models tailored to each client’s unique data distribution. Consequently, FedVLM can empower resource-constrained entities to leverage cutting-edge VLMs without incurring prohibitive computational costs, ensuring scalability and competitiveness in real-world applications.  

\vspace{2mm}
\noindent
\textbf{Rationale for Pre-trained Models and Personalized LoRA}:
Transformer-based models are highly adaptable across domains but are computationally expensive to train from scratch.  Fine-tuning pre-trained VLMs offers significant benefits, including faster convergence, improved generalization, zero-shot capabilities, and reduced resource consumption~\cite{li2023blip}. 
To optimize personalization in FL, we fine-tune a pre-trained VLM for VQA while maintaining efficiency. Given the resource constraints in FL, we employ PEFT methods to reduce computational overhead. Specifically, we aggregate only LoRA’s matrix $B$ while personalizing matrix $A$, keeping other model components frozen. This strategy ensures a balance between efficiency and effective adaptation, enabling client-specific fine-tuning without compromising scalability.

\vspace{2mm}
\noindent
\textbf{Limitations}: While our proposed FedVLM framework shows promising performance, it also inherits certain limitations common to federated and large vision-language models. First, personalized models may amplify biases present in localized client data, potentially compromising fairness and generalizability. Second, as with most large-scale generative models, there remains a risk of producing inaccurate or biased outputs. Additionally, without appropriate safeguards, training VLMs can lead to privacy leakage, as models may memorize sensitive image-text pairs~\cite{caldarella2024phantom}. Although privacy-preserving mechanisms are beyond the scope of this study, our focus is on demonstrating the feasibility of federated fine-tuning for VLMs. Addressing these concerns—bias mitigation, fairness, and privacy—remains essential for future work in this space.

%% file: 5_conclusion.tex
\section{Conclusion}
In this paper, we present FedVLM, a novel framework for scalable and privacy-aware vision-language model adaptation using federated learning. By integrating our personalized LoRA variant (pLoRA), FedVLM balances communication efficiency with client-specific fine-tuning. Experimental results show that FedVLM outperforms state-of-the-art methods, improving accuracy by 24.5\% in non-iid settings and 15.1\% in iid scenarios. These gains highlight FedVLM’s effectiveness in handling diverse data distributions while maintaining scalability. Although our implementation focuses on Florence-2, applying FedVLM to other VLM architectures is a promising direction for future research.

%% file: main.bbl
\begin{thebibliography}{49}
\providecommand{\natexlab}[1]{#1}
\providecommand{\url}[1]{\texttt{#1}}
\expandafter\ifx\csname urlstyle\endcsname\relax
  \providecommand{\doi}[1]{doi: #1}\else
  \providecommand{\doi}{doi: \begingroup \urlstyle{rm}\Url}\fi

\bibitem[Agbaje et~al.(2023)Agbaje, Anjum, Talukder, Islam, Nwafor, and Olufowobi]{agbaje2023fedcime}
P.~Agbaje, A.~Anjum, Z.~Talukder, M.~Islam, E.~Nwafor, and H.~Olufowobi.
\newblock Fedcime: An efficient federated learning approach for clients in mobile edge computing.
\newblock In \emph{2023 IEEE International Conference on Edge Computing and Communications (EDGE)}, pages 215--220. IEEE, 2023.

\bibitem[Anjum et~al.(2024)Anjum, Eren, Boureima, Alexandrov, and Bhattarai]{anjum2024tensor}
A.~Anjum, M.~E. Eren, I.~Boureima, B.~Alexandrov, and M.~Bhattarai.
\newblock Tensor train low-rank approximation (tt-lora): Democratizing ai with accelerated llms.
\newblock \emph{arXiv preprint arXiv:2408.01008}, 2024.

\bibitem[Caldarella et~al.(2024)Caldarella, Mancini, Ricci, and Aljundi]{caldarella2024phantom}
S.~Caldarella, M.~Mancini, E.~Ricci, and R.~Aljundi.
\newblock The phantom menace: unmasking privacy leakages in vision-language models.
\newblock \emph{arXiv preprint arXiv:2408.01228}, 2024.

\bibitem[Cho et~al.(2024)Cho, Liu, Xu, Fahrezi, and Joshi]{cho2024heterogeneous}
Y.~J. Cho, L.~Liu, Z.~Xu, A.~Fahrezi, and G.~Joshi.
\newblock Heterogeneous low-rank approximation for federated fine-tuning of on-device foundation models.
\newblock \emph{arXiv preprint arXiv:2401.06432}, 2024.

\bibitem[Ding et~al.(2022)Ding, Xiao, Codella, Luo, Wang, and Yuan]{ding2022davit}
M.~Ding, B.~Xiao, N.~Codella, P.~Luo, J.~Wang, and L.~Yuan.
\newblock Davit: Dual attention vision transformers.
\newblock In \emph{European conference on computer vision}, pages 74--92. Springer, 2022.

\bibitem[Goyal et~al.(2017)Goyal, Khot, Summers-Stay, Batra, and Parikh]{goyal2017making}
Y.~Goyal, T.~Khot, D.~Summers-Stay, D.~Batra, and D.~Parikh.
\newblock Making the v in vqa matter: Elevating the role of image understanding in visual question answering.
\newblock In \emph{Proceedings of the IEEE conference on computer vision and pattern recognition}, pages 6904--6913, 2017.

\bibitem[Han et~al.(2015)Han, Mao, and Dally]{han2015deep}
S.~Han, H.~Mao, and W.~J. Dally.
\newblock Deep compression: Compressing deep neural networks with pruning, trained quantization and huffman coding.
\newblock \emph{arXiv preprint arXiv:1510.00149}, 2015.

\bibitem[Hayou et~al.(2024)Hayou, Ghosh, and Yu]{hayou2024impact}
S.~Hayou, N.~Ghosh, and B.~Yu.
\newblock The impact of initialization on lora finetuning dynamics.
\newblock \emph{Advances in Neural Information Processing Systems}, 37:\penalty0 117015--117040, 2024.

\bibitem[Hinton(2015)]{hinton2015distilling}
G.~Hinton.
\newblock Distilling the knowledge in a neural network.
\newblock \emph{arXiv preprint arXiv:1503.02531}, 2015.

\bibitem[Howard et~al.(2019)Howard, Sandler, Chu, Chen, Chen, Tan, Wang, Zhu, Pang, Vasudevan, et~al.]{howard2019searching}
A.~Howard, M.~Sandler, G.~Chu, L.-C. Chen, B.~Chen, M.~Tan, W.~Wang, Y.~Zhu, R.~Pang, V.~Vasudevan, et~al.
\newblock Searching for mobilenetv3.
\newblock In \emph{Proceedings of the IEEE/CVF international conference on computer vision}, pages 1314--1324, 2019.

\bibitem[Hu et~al.(2023)Hu, Russell, Yeo, Murez, Fedoseev, Kendall, Shotton, and Corrado]{hu2023gaia}
A.~Hu, L.~Russell, H.~Yeo, Z.~Murez, G.~Fedoseev, A.~Kendall, J.~Shotton, and G.~Corrado.
\newblock Gaia-1: A generative world model for autonomous driving.
\newblock \emph{arXiv preprint arXiv:2309.17080}, 2023.

\bibitem[Hu et~al.(2021)Hu, Shen, Wallis, Allen-Zhu, Li, Wang, Wang, and Chen]{hu2021lora}
E.~J. Hu, Y.~Shen, P.~Wallis, Z.~Allen-Zhu, Y.~Li, S.~Wang, L.~Wang, and W.~Chen.
\newblock Lora: Low-rank adaptation of large language models.
\newblock \emph{arXiv preprint arXiv:2106.09685}, 2021.

\bibitem[Huang et~al.(2023)Huang, Bianchi, Yuksekgonul, Montine, and Zou]{huang2023visual}
Z.~Huang, F.~Bianchi, M.~Yuksekgonul, T.~J. Montine, and J.~Zou.
\newblock A visual--language foundation model for pathology image analysis using medical twitter.
\newblock \emph{Nature medicine}, 29\penalty0 (9):\penalty0 2307--2316, 2023.

\bibitem[Hurst et~al.(2024)Hurst, Lerer, Goucher, Perelman, Ramesh, Clark, Ostrow, Welihinda, Hayes, Radford, et~al.]{hurst2024gpt}
A.~Hurst, A.~Lerer, A.~P. Goucher, A.~Perelman, A.~Ramesh, A.~Clark, A.~Ostrow, A.~Welihinda, A.~Hayes, A.~Radford, et~al.
\newblock Gpt-4o system card.
\newblock \emph{arXiv preprint arXiv:2410.21276}, 2024.

\bibitem[Jacob et~al.(2018)Jacob, Kligys, Chen, Zhu, Tang, Howard, Adam, and Kalenichenko]{jacob2018quantization}
B.~Jacob, S.~Kligys, B.~Chen, M.~Zhu, M.~Tang, A.~Howard, H.~Adam, and D.~Kalenichenko.
\newblock Quantization and training of neural networks for efficient integer-arithmetic-only inference.
\newblock In \emph{Proceedings of the IEEE conference on computer vision and pattern recognition}, pages 2704--2713, 2018.

\bibitem[Kemsaram et~al.(2019)Kemsaram, Das, and Dubbelman]{kemsaram2019integrated}
N.~Kemsaram, A.~Das, and G.~Dubbelman.
\newblock An integrated framework for autonomous driving: Object detection, lane detection, and free space detection.
\newblock In \emph{2019 Third World Conference on Smart Trends in Systems Security and Sustainablity (WorldS4)}, pages 260--265. IEEE, 2019.

\bibitem[Krizhevsky et~al.(2009)Krizhevsky, Hinton, et~al.]{krizhevsky2009learning}
A.~Krizhevsky, G.~Hinton, et~al.
\newblock Learning multiple layers of features from tiny images.
\newblock 2009.

\bibitem[Le et~al.(2023)Le, Nguyen, Thwal, Qiao, Zhang, and Hong]{le2023fedmekt}
H.~Q. Le, M.~N. Nguyen, C.~M. Thwal, Y.~Qiao, C.~Zhang, and C.~S. Hong.
\newblock Fedmekt: Distillation-based embedding knowledge transfer for multimodal federated learning.
\newblock \emph{arXiv preprint arXiv:2307.13214}, 2023.

\bibitem[Lewis(2019)]{lewis2019bart}
M.~Lewis.
\newblock Bart: Denoising sequence-to-sequence pre-training for natural language generation, translation, and comprehension.
\newblock \emph{arXiv preprint arXiv:1910.13461}, 2019.

\bibitem[Li et~al.(2021)Li, Selvaraju, Gotmare, Joty, Xiong, and Hoi]{li2021align}
J.~Li, R.~Selvaraju, A.~Gotmare, S.~Joty, C.~Xiong, and S.~C.~H. Hoi.
\newblock Align before fuse: Vision and language representation learning with momentum distillation.
\newblock \emph{Advances in neural information processing systems}, 34:\penalty0 9694--9705, 2021.

\bibitem[Li et~al.(2022)Li, Li, Xiong, and Hoi]{li2022blip}
J.~Li, D.~Li, C.~Xiong, and S.~Hoi.
\newblock Blip: Bootstrapping language-image pre-training for unified vision-language understanding and generation.
\newblock In \emph{International conference on machine learning}, pages 12888--12900. PMLR, 2022.

\bibitem[Li et~al.(2023{\natexlab{a}})Li, Li, Savarese, and Hoi]{li2023blip}
J.~Li, D.~Li, S.~Savarese, and S.~Hoi.
\newblock Blip-2: Bootstrapping language-image pre-training with frozen image encoders and large language models.
\newblock In \emph{International conference on machine learning}, pages 19730--19742. PMLR, 2023{\natexlab{a}}.

\bibitem[Li et~al.(2020)Li, Sahu, Zaheer, Sanjabi, Talwalkar, and Smith]{li2020federated}
T.~Li, A.~K. Sahu, M.~Zaheer, M.~Sanjabi, A.~Talwalkar, and V.~Smith.
\newblock Federated optimization in heterogeneous networks.
\newblock \emph{Proceedings of Machine learning and systems}, 2:\penalty0 429--450, 2020.

\bibitem[Li et~al.(2024)Li, Wu, Jiang, Guo, Gong, Cao, Liu, Jiang, and Sun]{li2024enhancing}
X.~Li, Y.~Wu, X.~Jiang, Z.~Guo, M.~Gong, H.~Cao, Y.~Liu, D.~Jiang, and X.~Sun.
\newblock Enhancing visual document understanding with contrastive learning in large visual-language models.
\newblock In \emph{Proceedings of the IEEE/CVF Conference on Computer Vision and Pattern Recognition}, pages 15546--15555, 2024.

\bibitem[Li et~al.(2023{\natexlab{b}})Li, Bubeck, Eldan, Del~Giorno, Gunasekar, and Lee]{textbooks2}
Y.~Li, S.~Bubeck, R.~Eldan, A.~Del~Giorno, S.~Gunasekar, and Y.~T. Lee.
\newblock Textbooks are all you need ii: \textbf{phi-1.5} technical report.
\newblock \emph{arXiv preprint arXiv:2309.05463}, 2023{\natexlab{b}}.

\bibitem[Lin et~al.(2023)Lin, Yin, Ping, Lu, Molchanov, Tao, Mao, Kautz, Shoeybi, and Han]{lin2023vila}
J.~Lin, H.~Yin, W.~Ping, Y.~Lu, P.~Molchanov, A.~Tao, H.~Mao, J.~Kautz, M.~Shoeybi, and S.~Han.
\newblock Vila: On pre-training for visual language models.
\newblock \emph{arXiv preprint arXiv:2312.07533}, 2023.

\bibitem[Lin et~al.(2014)Lin, Maire, Belongie, Hays, Perona, Ramanan, Doll{\'a}r, and Zitnick]{lin2014microsoft}
T.-Y. Lin, M.~Maire, S.~Belongie, J.~Hays, P.~Perona, D.~Ramanan, P.~Doll{\'a}r, and C.~L. Zitnick.
\newblock Microsoft coco: Common objects in context.
\newblock In \emph{Computer Vision--ECCV 2014: 13th European Conference, Zurich, Switzerland, September 6-12, 2014, Proceedings, Part V 13}, pages 740--755. Springer, 2014.

\bibitem[Liu et~al.(2024)Liu, Zhao, Iandola, Lai, Tian, Fedorov, Xiong, Chang, Shi, Krishnamoorthi, et~al.]{liu2024mobilellm}
Z.~Liu, C.~Zhao, F.~Iandola, C.~Lai, Y.~Tian, I.~Fedorov, Y.~Xiong, E.~Chang, Y.~Shi, R.~Krishnamoorthi, et~al.
\newblock Mobilellm: Optimizing sub-billion parameter language models for on-device use cases.
\newblock \emph{arXiv preprint arXiv:2402.14905}, 2024.

\bibitem[Marino et~al.(2019)Marino, Rastegari, Farhadi, and Mottaghi]{marino2019ok}
K.~Marino, M.~Rastegari, A.~Farhadi, and R.~Mottaghi.
\newblock Ok-vqa: A visual question answering benchmark requiring external knowledge.
\newblock In \emph{Proceedings of the IEEE/cvf conference on computer vision and pattern recognition}, pages 3195--3204, 2019.

\bibitem[McMahan et~al.(2017)McMahan, Moore, Ramage, Hampson, and y~Arcas]{mcmahan2017communication}
B.~McMahan, E.~Moore, D.~Ramage, S.~Hampson, and B.~A. y~Arcas.
\newblock Communication-efficient learning of deep networks from decentralized data.
\newblock In \emph{Artificial intelligence and statistics}, pages 1273--1282. PMLR, 2017.

\bibitem[Nguyen et~al.(2024)Nguyen, Munoz, and Jannesari]{nguyen2024flora}
D.~P. Nguyen, J.~P. Munoz, and A.~Jannesari.
\newblock Flora: Enhancing vision-language models with parameter-efficient federated learning.
\newblock \emph{arXiv preprint arXiv:2404.15182}, 2024.

\bibitem[Qi et~al.(2024)Qi, Luan, Huang, Fung, Yang, and Qian]{qi2024fdlora}
J.~Qi, Z.~Luan, S.~Huang, C.~Fung, H.~Yang, and D.~Qian.
\newblock Fdlora: Personalized federated learning of large language model via dual lora tuning.
\newblock \emph{arXiv preprint arXiv:2406.07925}, 2024.

\bibitem[Radford et~al.(2021)Radford, Kim, Hallacy, Ramesh, Goh, Agarwal, Sastry, Askell, Mishkin, Clark, et~al.]{radford2021learning}
A.~Radford, J.~W. Kim, C.~Hallacy, A.~Ramesh, G.~Goh, S.~Agarwal, G.~Sastry, A.~Askell, P.~Mishkin, J.~Clark, et~al.
\newblock Learning transferable visual models from natural language supervision.
\newblock In \emph{International conference on machine learning}, pages 8748--8763. PMLR, 2021.

\bibitem[Saha et~al.(2024)Saha, Van~Horn, and Maji]{saha2024improved}
O.~Saha, G.~Van~Horn, and S.~Maji.
\newblock Improved zero-shot classification by adapting vlms with text descriptions.
\newblock In \emph{Proceedings of the IEEE/CVF Conference on Computer Vision and Pattern Recognition}, pages 17542--17552, 2024.

\bibitem[Sima et~al.(2023)Sima, Renz, Chitta, Chen, Zhang, Xie, Luo, Geiger, and Li]{sima2023drivelm}
C.~Sima, K.~Renz, K.~Chitta, L.~Chen, H.~Zhang, C.~Xie, P.~Luo, A.~Geiger, and H.~Li.
\newblock Drivelm: Driving with graph visual question answering.
\newblock \emph{arXiv preprint arXiv:2312.14150}, 2023.

\bibitem[Sun et~al.(2024)Sun, Li, Li, and Ding]{sun2024improving}
Y.~Sun, Z.~Li, Y.~Li, and B.~Ding.
\newblock Improving lora in privacy-preserving federated learning.
\newblock \emph{arXiv preprint arXiv:2403.12313}, 2024.

\bibitem[Tan et~al.(2022)Tan, Long, Ma, Liu, Zhou, and Jiang]{tan2022federated}
Y.~Tan, G.~Long, J.~Ma, L.~Liu, T.~Zhou, and J.~Jiang.
\newblock Federated learning from pre-trained models: A contrastive learning approach.
\newblock \emph{Advances in neural information processing systems}, 35:\penalty0 19332--19344, 2022.

\bibitem[Team(2024)]{team2024chameleon}
C.~Team.
\newblock Chameleon: Mixed-modal early-fusion foundation models.
\newblock \emph{arXiv preprint arXiv:2405.09818}, 2024.

\bibitem[Vaswani(2017)]{vaswani2017attention}
A.~Vaswani.
\newblock Attention is all you need.
\newblock \emph{Advances in Neural Information Processing Systems}, 2017.

\bibitem[Wang et~al.(2024)Wang, Zheng, Han, Xu, Li, and Zhang]{wang2024fednlr}
H.~Wang, P.~Zheng, X.~Han, W.~Xu, R.~Li, and T.~Zhang.
\newblock Fednlr: Federated learning with neuron-wise learning rates.
\newblock In \emph{Proceedings of the 30th ACM SIGKDD Conference on Knowledge Discovery and Data Mining}, pages 3069--3080, 2024.

\bibitem[Wu et~al.(2024)Wu, Li, Li, Ding, and Gao]{wu2024fedbiot}
F.~Wu, Z.~Li, Y.~Li, B.~Ding, and J.~Gao.
\newblock Fedbiot: Llm local fine-tuning in federated learning without full model.
\newblock In \emph{Proceedings of the 30th ACM SIGKDD Conference on Knowledge Discovery and Data Mining}, pages 3345--3355, 2024.

\bibitem[Xiao et~al.(2024)Xiao, Wu, Xu, Dai, Hu, Lu, Zeng, Liu, and Yuan]{xiao2024florence}
B.~Xiao, H.~Wu, W.~Xu, X.~Dai, H.~Hu, Y.~Lu, M.~Zeng, C.~Liu, and L.~Yuan.
\newblock Florence-2: Advancing a unified representation for a variety of vision tasks.
\newblock In \emph{Proceedings of the IEEE/CVF Conference on Computer Vision and Pattern Recognition}, pages 4818--4829, 2024.

\bibitem[Yu et~al.(2023)Yu, Liu, Wang, Xu, and Liu]{yu2023multimodal}
Q.~Yu, Y.~Liu, Y.~Wang, K.~Xu, and J.~Liu.
\newblock Multimodal federated learning via contrastive representation ensemble.
\newblock \emph{arXiv preprint arXiv:2302.08888}, 2023.

\bibitem[Yu and Li(2021)]{yu2021toward}
R.~Yu and P.~Li.
\newblock Toward resource-efficient federated learning in mobile edge computing.
\newblock \emph{IEEE Network}, 35\penalty0 (1):\penalty0 148--155, 2021.

\bibitem[Yu et~al.(2024)Yu, Zhang, Yao, Dang, Chen, Lu, Cui, He, Liu, Chua, et~al.]{yu2024rlaif}
T.~Yu, H.~Zhang, Y.~Yao, Y.~Dang, D.~Chen, X.~Lu, G.~Cui, T.~He, Z.~Liu, T.-S. Chua, et~al.
\newblock Rlaif-v: Aligning mllms through open-source ai feedback for super gpt-4v trustworthiness.
\newblock \emph{arXiv preprint arXiv:2405.17220}, 2024.

\bibitem[Zhang et~al.(2021)Zhang, Guo, Ma, Wang, Xu, and Wu]{zhang2021parameterized}
J.~Zhang, S.~Guo, X.~Ma, H.~Wang, W.~Xu, and F.~Wu.
\newblock Parameterized knowledge transfer for personalized federated learning.
\newblock \emph{Advances in Neural Information Processing Systems}, 34:\penalty0 10092--10104, 2021.

\bibitem[Zhao et~al.(2022)Zhao, Barnaghi, and Haddadi]{zhao2022multimodal}
Y.~Zhao, P.~Barnaghi, and H.~Haddadi.
\newblock Multimodal federated learning on iot data.
\newblock In \emph{2022 IEEE/ACM Seventh International Conference on Internet-of-Things Design and Implementation (IoTDI)}, pages 43--54. IEEE, 2022.

\bibitem[Zhi et~al.(2024)Zhi, Bi, Xu, Wang, and Xiang]{zhi2024knowledge}
M.~Zhi, Y.~Bi, W.~Xu, H.~Wang, and T.~Xiang.
\newblock Knowledge-aware parameter coaching for personalized federated learning.
\newblock In \emph{Proceedings of the AAAI Conference on Artificial Intelligence}, volume~38, pages 17069--17077, 2024.

\bibitem[Zhou et~al.(2024)Zhou, Hu, Weng, Jia, Luo, Liu, Wu, and Huang]{zhou2024tinyllava}
B.~Zhou, Y.~Hu, X.~Weng, J.~Jia, J.~Luo, X.~Liu, J.~Wu, and L.~Huang.
\newblock Tinyllava: A framework of small-scale large multimodal models.
\newblock \emph{arXiv preprint arXiv:2402.14289}, 2024.

\end{thebibliography}
